\documentclass[a4paper,fleqn]{cas-sc}

\usepackage[authoryear,longnamesfirst]{natbib}
\usepackage{gensymb}

\def\tsc#1{\csdef{#1}{\textsc{\lowercase{#1}}\xspace}}
\tsc{WGM}
\tsc{QE}

\begin{document}
\let\WriteBookmarks\relax
\def\floatpagepagefraction{1}
\def\textpagefraction{.001}

\shorttitle{}    

\shortauthors{}  

\title [mode = title]{Variational LSTM with Augmented Inputs: Nonlinear Response History Metamodeling with Aleatoric and Epistemic  Uncertainty}  



%

\author[1]{Manisha Sapkota}


\fnmark[1]

\ead{manisapk@ttu.edu}

\ead[url]{}


\affiliation[1]{organization={Department of Civil, Environmental, and Construction Engineering, Texas Tech University},
            addressline={2500 Broadway St}, 
            city={Lubbock},
          citysep={}, 
            postcode={79409}, 
            state={TX},
            country={USA}}

\author[2]{Min Li}

\fnmark[2]

\ead{lim33@rpi.edu}

\ead[url]{https://faculty.rpi.edu/min-li}


\affiliation[2]{organization={Department of Civil and Environmental Engineering, Rensselaer Polytechnic Institute},
            addressline={110 8th Street}, 
            city={Troy},
            citysep={}, 
            postcode={12180}, 
            state={NY},
            country={USA}}
            
\author[1,3]{Bowei Li}[orcid=0000-0002-6859-4939]
\cormark[1]

\fnmark[3]

\ead{boweili@ttu.edu}

\ead[url]{https://sites.google.com/view/boweiliphd/home}



\affiliation[3]{organization={National Wind Institute, Texas Tech University},
            addressline={1009 Canton Ave}, 
            city={Lubbock},
            citysep={}, 
            postcode={79409}, 
            state={TX},
            country={USA}}
\cortext[1]{Corresponding author}



\begin{abstract}
Uncertainty propagation in high-dimensional nonlinear dynamic structural systems is pivotal in state-of-the-art performance-based design and risk assessment, where uncertainties from both excitations and structures, i.e., the aleatoric uncertainty, must be considered. This poses a significant challenge due to heavy computational demands. Machine learning techniques are thus introduced as metamodels to alleviate this burden. However, the "black box" nature of Machine learning models underscores the necessity of avoiding overly confident predictions, particularly when data and training efforts are insufficient. This creates a need, in addition to considering the aleatoric uncertainty, of estimating the uncertainty related to the prediction confidence, i.e., epistemic uncertainty, for machine learning-based metamodels. We developed a probabilistic metamodeling technique based on a variational long short-term memory (LSTM) with augmented inputs to simultaneously capture aleatoric and epistemic uncertainties. Key random system parameters are treated as augmented inputs alongside excitation series carrying record-to-record variability to capture the full range of aleatoric uncertainty. Meanwhile, epistemic uncertainty is effectively approximated via the Monte Carlo dropout scheme. Unlike computationally expensive full Bayesian approaches, this method incurs negligible additional training costs while enabling nearly cost-free uncertainty simulation. The proposed technique is demonstrated through multiple case studies involving stochastic seismic or wind excitations. Results show that the calibrated metamodels accurately reproduce nonlinear response time histories and provide confidence bounds indicating the associated epistemic uncertainty.
\end{abstract}


\begin{highlights}
\item Variational LSTM with augmented inputs for nonlinear structural metamodeling.
\item Augmented inputs capture record-to-record variability and system uncertainty.
\item Epistemic uncertainty in response histories is quantified via Monte Carlo dropout.
\item Validated on strongly nonlinear systems under stochastic seismic and wind loads.
\item Monte Carlo dropout reflects prediction fidelity across diverse scenarios.
\end{highlights}


\maketitle

\section{Introduction}\label{Sec:Intro}
The state-of-the-art performance-based design and risk assessment have gained profound interest and significant advancement in recent years \cite{moehle2004framework,baker2008uncertainty,augusti2008performance,yang2009seismic,barbato2013performance,ciampoli2011performance,gunay2013peer,gardoni2016multi}. This is motivated by the fundamental need of rational engineering design and decision through comprehensive and accurate characterization of structural performances under hazards, e.g., earthquake, wind, etc. To this end, explicit structural dynamic analysis under a full range of hazard intensities, including those induce nonlinearity \cite{hong2004accumulation,bezabeh2021nonlinear,huang2022inelastic}, is required to quantify structural performances under these hazards. Moreover, practical engineered structural systems are in general involving a great number of degree-of-freedoms (DoFs), making the structural analysis a high-dimensional nonlinear dynamic analysis problem, a well-known computationally demanding task. In addition, this simulation task in nature involves uncertainty propagation because of the ubiquitous uncertainty present in both excitations and structural systems \cite{ellingwood2008structural}. This requires a massive number of repeated expensive high-dimensional nonlinear dynamic analyses for Monte Carlo simulation \cite{vamvatsikos2002incremental,seo2013estimating,spence2014performance,judd2015inelastic,bernardini2015performance,chuang2017performance,cui2018unified,ierimonti2019cost,micheli2019performance,cui2020performance,ouyang2020performance}, leading to an impractical task due to the unaffordable computational demand. This is now the major barrier to advance the engineering adoption of existing state-of-the-art design and assessment frameworks.

A potential remedy to address the computational demand is metamodeling, establishing an accurate but computationally tractable surrogate for solving structural dynamic responses. Response surface approach is introduced for this purpose by taking intensity (peak ground acceleration, wind speed, etc.) and system parameters (material properties, structural damping, etc.) as inputs to reproduce response indices (interstory drift ratio, etc.) \cite{towashiraporn2004building,vaidyanathan2005artificial,de2009numerical,seo2012metamodel,perotti2013numerical,saha2016uncertainty,kroetz2017performance,segura2020metamodel}. In these implementations, response surfaces are constructed by statistical and machine learning approaches by fitting the input-output mapping. Since propagating high-dimensional record-to-record variability directly through these mappings is challenging, a dual response surface approach is adopted to simultaneously approximate the means and reproduce the standard deviations of the responses. The response surface approach is shown to be effective in various engineering applications and has thus been widely used. However, this is usually insufficient to accurately represent the nonstationary and higher-order statistics of nonlinear structural responses. To address this issue, it is required to explicitly propagate the record-to-record variability. To this end, nonlinear autoregressive models with exogenous inputs (NARX) \cite{spiridonakos2015metamodeling,mai2016surrogate,Bhattacharyya20,li2021response}, a metamodeling technique with sequence processing capabilities, is introduced. The NARX model constitutes a recursive mapping that explicitly captures the high-dimensional record-to-record variability by incorporating the system excitation as a step-by-step input. On the other hand, more versatile deep learning architectures, including convolutional neural networks (CNN) \cite{kim2019response,zhang2020physicscnn}, long short-term memory (LSTM) \cite{kundu2020deep,wang2020knowledge,zhang2020physics,simpson2021machine,li2022metamodeling}, and neural operators \cite{GOSWAMI2025121284}, were reported to be applied in reconstructing response time histories. Like NARX, these models are also sequence-to-sequence mappings that explicitly account for record-to-record variability propagation. To allow these models to be used in practical high-dimensional structural systems, dimensionality reduction through a proper orthogonal basis or an autoencoder is introduced, enabling deep learning techniques to be effectively applied to full-scale engineering structures that have a great number of degrees-of-freedom (DoFs) \cite{li2021response,simpson2021machine,li2022metamodeling}. 

Nevertheless, these deep learning architectures only capture record-to-record variability, representing just one component of the full range of uncertainties that must be accounted for. In general, the uncertainties include aleatoric uncertainty, irreducible randomness inherent in the problem setting, and epistemic uncertainty, which quantifies model confidence and can be reduced through additional data or calibration \cite{der2009aleatory,kendall2017uncertainties}. The record-to-record variability belongs to aleatoric uncertainty. In addition, system uncertainties, e.g., material properties and system properties, constitute an important component of aleatoric uncertainty; however, they are not captured by these aforementioned deep learning architectures. Although in NARX, polynomial chaos expansions \cite{spiridonakos2015metamodeling,mai2016surrogate} and Kriging \cite{Bhattacharyya20} were leveraged to identify the relationship between system parameters and NARX coefficients, extending these approaches to deep neural networks remains challenging due to their high-dimensional spaces of trainable parameters and complex architectures. A potential solution is to leverage generative schemes or probabilistic machine learning techniques that aim at capturing the probability distribution of responses, instead of individual samples \cite{xiong2025uncertainty,peng2026accelerating}. However, when the interpretation of system performance in terms of individual samples is demanded, explicit correspondence from excitation and system realizations to responses is usually needed. To establish this correspondence, an alternative is to include system parameters as inputs to fuse into neural network features, e.g., through a network branch \cite{li2024time,ning2024dual}. As such, the neural network learns to capture a one-to-one mapping from system parameters to responses.

The importance of epistemic uncertainty cannot be overstated, considering the potential catastrophic consequences if using weak response predictions with overconfidence in design and risk tasks where safety is critical \cite{gal2016uncertainty}. This is particularly true for machine learning techniques, of which the black box nature makes it challenging to guarantee their performance. In existing metamodeling approaches, Kriging \cite{chatterjee2017efficient,gidaris2015kriging,ghosh2019kriging,shi2019adaptive,kyprioti2021kriging} offers a straightforward means of quantifying epistemic uncertainty; however, it fails in capturing high-dimensional sequence-to-sequence relationships. A potential solution is to introduce Bayesian machine learning approaches \cite{kim2020probabilistic,ghosh2020seismic,ding2023efficient} to the aforementioned sequence-to-sequence mapping architectures, where the epistemic uncertainty is quantified as probabilistic prediction based on posterior weight distributions inferred from available data. However, for large machine learning models, analytical posterior weight distributions are in general not available due to the nonconjugacy and nonlinearity. On the other hand, commonly used Markov Chain Monte Carlo (MCMC) \cite{straub2008improved,koutsourelakis2010assessing,pujari2016bayesian,xu2016probabilistic} are computationally very demanding due to the great number of trainable parameters involved in large machine learning models. Bootstrap \cite{tibshirani1993introduction}, a Bayesian method through sampling multiple datasets by adding noise/disturbance and training an estimator for each of the disturbed datasets, suffers from a significant increase in computational demand in training. An alternative to quantifying epistemic uncertainty efficiently is Monte Carlo dropout, a variational inference approach by approximating the posterior weight distribution through randomly masking out hidden neurons \cite{gal2015dropout}. This Monte Carlo dropout can be directly embedded in existing training algorithms \cite{hinton2012improving,srivastava2014dropout} with minimal increase in computational demand, making it an efficient approach allowing large machine learning models to quantify epistemic uncertainty.


In this paper, a variational LSTM with augmented inputs metamodeling approach is developed for nonlinear dynamic structural systems, with the capability of modeling both aleatoric (record-to-record variability and system uncertainty) and epistemic uncertainty. Proper orthogonal decomposition (POD)-based dimensionality reduction and wavelet approximation are first considered to address the high-dimensionality of structural systems and long duration of excitations and responses. The LSTM network is leveraged as a sequence-to-sequence mapping approach to capture all aleatoric uncertainty, including record-to-record variability in excitation series and parametric system uncertainty. In particular, the parametric system uncertainty is considered by augmenting its realizations to the network input series. In addition, epistemic uncertainty is modeled by implementing the variational inference through the Monte Carlo dropout for the LSTM network. Once calibrated, the LSTM network offers an efficient one-to-one mapping from stochastic excitation and system parameters to a full range of response time histories, while allowing for convenient estimation of the remaining epistemic uncertainty through the Monte Carlo dropout in inference. The applicability of the proposed metamodeling technique is demonstrated by multiple case studies involving nonlinear dynamic structural systems with system uncertainty while subjected to stochastic seismic or wind excitations. 

\section{Problem setting}\label{Sec:problem}
In general, dynamics of a structural system with $n_\text{{DoF}}$ DoFs subjected to stochastic excitation $\boldsymbol{F}(t)$ can be described by the following equilibrium over a time horizon of $t\in\mathcal{T}$:
\begin{equation}
\label{eq1}
\mathcal{F}(\ddot{\boldsymbol{U}}(t),\dot{\boldsymbol{U}}(t),\boldsymbol{U}(t);\mathbf{\Theta})=\boldsymbol{F}(t;\textbf{R}),~t\in\mathcal{T}
\end{equation}
where $\mathcal{F}$ is the operator that represents the resultant internal forces, including inertia, damping, and restoring forces; $\mathbf{\Theta}\in(\Omega_{\Theta}, \mathcal{B}_{\Theta}, \mathcal{P}_{\Theta})$ is a random vector of size $n_{\Theta}$ characterizing the system uncertainty, defined by its sample space $\Omega_{\Theta}=\mathbb{R}^{n_{\Theta}}$, sigma-algebra $\mathcal{B}_{\Theta}$, and probability measure $\mathcal{P}_{\Theta}$; $\textbf{R}\in(\Omega_{\textbf{R}}, \mathcal{B}_{\textbf{R}}, \mathcal{P}_{\textbf{R}})$ is a random process with sample space $\Omega_{\textbf{R}}=\mathbb{R}^{n_\text{{DoF}}}\times\mathcal{T}$, sigma-algebra $\mathcal{B}_{\textbf{R}}$, and probability measure $\mathcal{P}_{\textbf{R}}$ that defines the record-to-record variability (stochasticity) of the excitation; $\ddot{\boldsymbol{U}}(t),\dot{\boldsymbol{U}}(t),\boldsymbol{U}(t)$ are respectively the acceleration, velocity, and displacement, all are random processes derived from $\mathbf{\Theta}$ and $\textbf{R}$, and with sample spaces of $\mathbb{R}^{n_\text{{DoF}}}\times\mathcal{T}$. When considering a full spectrum of excitation intensity due to the variability in hazard characteristics, $\mathcal{F}$ can exhibit complex nonlinear behaviors that, in most cases, exclude alternative solutions but the time-consuming direct integration schemes. This leads to a computationally very challenging nonlinear dynamic problem defined over a potentially long time horizon and with high-dimensionality in equations and randomness. 

The sequence-to-sequence mapping architectures, including LSTM, CNN, and DeepONet, aforementioned in the introduction, offer a remedy for this high computational demand by providing a highly efficient sequence-to-sequence mapping from excitations to response time histories. This allows the explicit propagation of the high-dimensional record-to-record variability. However, the record-to-record variability is only a subset of the aleatoric uncertainty. System uncertainty, an additional element of aleatoric uncertainty, remains outside the current scope of this analysis. In contrast to record-to-record variability, which is directly propagated through excitations, system uncertainties are difficult to integrate into the neural networks due to their 'black box' architecture and the massive number of trainable parameters. Additionally, unlike traditional numerical solvers, these networks often lack guaranteed physical validity. Thus, it is particularly necessary to avoid overly confident predictions if used in safety-related engineering tasks, e.g., performance or risk assessments. This requires quantifying the uncertainty related to the lack of knowledge in the neural networks, the so-called epistemic uncertainty. However, the complex and nonlinear architecture of the neural networks excludes a straightforward quantification of this epistemic uncertainty by either analytical or direct numerical simulations. These limitations necessitate the development of an efficient sequence-to-sequence mapping $\mathcal{M}(\cdot)$ that effectively captures both aleatoric and epistemic uncertainty, reads:
\begin{equation}
    \hat{\boldsymbol{U}}(t)=\mathcal{M}(\boldsymbol{F}(t;\textbf{R}), \mathbf{\Theta}; \mathbf{W})
\end{equation}
where $\hat{\boldsymbol{U}}(t)$ is the predicted stochastic displacement time history with its randomness induced by the aleatoric and epistemic uncertainty; $\mathbf{W}$ collects random trainable parameters of $\mathcal{M}(\cdot)$, containing randomness modeling the epistemic uncertainty of the mapping $\mathcal{M}(\cdot)$ to be propagated to $\hat{\boldsymbol{U}}(t)$.

To this end, we proposed a metamodeling approach based on variational LSTM with augmented inputs. This metamodeling approach includes system uncertainty as augmented inputs, in addition to the original inputs carrying the record-to-record variability, allowing explicit one-to-one mapping from the full range of aleatoric uncertainty to responses. The variational inference through the Monte Carlo dropout was considered to enable a computationally efficient means to estimate the epistemic uncertainty reflecting finite data and training effort. As such, this metamodeling technique offers an efficient, easy-to-implement approach to propagating both aleatoric and epistemic uncertainty for high-dimensional nonlinear dynamic systems.

\section{LSTM-based metamodeling}\label{Sec:augmented}
\subsection{LSTM network}\label{Sec:lstm}
The LSTM network is a type of recurrent neural network, which feeds neuron output as input. This recursive nature is suitable for time series data with underlying dynamics. An LSTM network is designed to significantly alleviate gradient vanishing and exploding issues that are typically seen in vanilla recurrent neural networks. This greatly enhances the applicability of the LSTM network in practice by allowing it to maintain exceptional stability in training and to capture long-term dependence in time series. Core to the LSTM network is the LSTM layer defined as:
\begin{align}
\label{eq:lstm}
    \textbf{g}_\text{f}(\tau) & = \sigma_\text{g}(\mathbf{w}_{\text{f},\text{H}}^\text{T}\textbf{y}(\tau-1)+\mathbf{w}_{\text{f},\text{I}}^\text{T}\textbf{x}(\tau)+\textbf{b}_{\text{f}})\\
    \textbf{g}_\text{i}(\tau) & = \sigma_\text{g}(\mathbf{w}_{\text{i},\text{H}}^\text{T}\textbf{y}(\tau-1)+\mathbf{w}_{\text{i},\text{I}}^\text{T}\textbf{x}(\tau)+\textbf{b}_{\text{i}})\\
    \textbf{g}_\text{o}(\tau) & = \sigma_\text{g}(\mathbf{w}_{\text{o},\text{H}}^\text{T}\textbf{y}(\tau-1)+\mathbf{w}_{\text{o},\text{I}}^\text{T}\textbf{x}(\tau)+\textbf{b}_{\text{o}})\\
    \Delta\textbf{s}(\tau) & = \sigma_\text{s}(\mathbf{w}_{\text{s},\text{H}}^\text{T}\textbf{y}(\tau-1)+\mathbf{w}_{\text{s},\text{I}}^\text{T}\textbf{x}(\tau)+\textbf{b}_{\text{s}})\\
    \textbf{s}(\tau) & = \textbf{g}_\text{f}(\tau)\circ\textbf{s}(\tau-1)+\textbf{g}_\text{i}(\tau)\circ\Delta\textbf{s}(\tau)\\
    \textbf{y}(\tau) & = \textbf{g}_\text{o}(\tau)\circ\sigma_\text{s}(\textbf{s}(\tau))
\end{align}
where $\tau$ represent the time index; $\textbf{x}$ and $\textbf{y}$ are the layer inputs and outputs; $\textbf{g}_\text{f}(\cdot)$, $\textbf{g}_\text{i}(\cdot)$, $\textbf{g}_\text{o}(\cdot)$ are respectively the forget, input, and output gates; $\sigma_\text{g}$ and $\sigma_\text{s}$ are gate and state activation functions taking sigmoid and hyperbolic tangent forms, respectively; $\textbf{s}$ and $\Delta\textbf{s}$ are cell state and its updating increment; $\circ$ indicates element-wise multiplication. This LSTM layer has trainable parameters including weights ($\mathbf{w}_{\text{f},\text{H}}$, $ \mathbf{w}_{\text{i},\text{H}}$, $\mathbf{w}_{\text{o},\text{H}}$, $\mathbf{w}_{\text{s},\text{H}}$, $\mathbf{w}_{\text{f},\text{I}}$, $\mathbf{w}_{\text{i},\text{I}}$, $\mathbf{w}_{\text{o},\text{I}}$, $\mathbf{w}_{\text{s},\text{I}}$) and bias ($\textbf{b}_{\text{f}}$, $\textbf{b}_{\text{i}}$, $\textbf{b}_{\text{o}}$, and $\textbf{b}_{\text{s}}$). In practice, an LSTM network is usually a neural network that consists of an input layer, LSTM layer(s), and a fully connected layer. The LSTM network is calibrated by minimizing a loss function $\mathcal{L}(\mathbf{w})$ with respect to all the trainable parameters $\mathbf{w}$. The loss function $\mathcal{L}(\mathbf{w})$ is typically defined as the discrepancy between the neural network outputs and ground truth data. Usually, this can be effectively solved by stochastic gradient descent or its variants. 

\subsection{Data processing}\label{Sec:process}
To calibrate the LSTM network, a dataset $\mathcal{D} = \{(\mathbf{f}_i, \boldsymbol{\theta}_i,\mathbf{u}_i), i=1,2,...,n\}$ containing $n$ realizations of stochastic excitation $\mathbf{F}$, structural systems $\mathbf{\Theta}$, and responses $\mathbf{U}$, is generated as described in Section~\ref{Sec:problem} \footnote{The notations follow standard conventions in probability theory, where random variables are denoted as capital and their realizations are represented by same letters but in lowercase. This convention applies to the remainder of this paper.}. The high dimensionality of the problem and the long time horizon make it intractable to directly use machine learning models for the dataset. To cope with the high dimensionality, projection-based dimensionality reduction is used:
\begin{align}
\label{pod_u} 
\mathbf{q}=\boldsymbol{\phi}^\text{T}\mathbf{u}\\
\label{pod_f}
\mathbf{p}=\boldsymbol{\phi}^\text{T}\mathbf{f}
\end{align}
with $\boldsymbol{\phi}\in\mathbb{R}^{n_\text{{DoF}}\times n_\text{r}}$ is truncated POD basis containing $n_\text{r}$ POD mode vectors within each of its columns. The truncated POD basis is obtained by performing singular value decomposition for snapshots collected from the response realizations $\boldsymbol{u}_i$ in $\mathcal{D}$. This gives in total $n_\text{{DoF}}$ POD mode vectors and those corresponding to the first few $n_\text{r}$ largest singular values are preserved as the truncated POD basis for the projection, ensuring the majority of response signal energy is captured. Through Eq.~(\ref{pod_u} and \ref{pod_f}), the high-dimensional excitation $\mathbf{f}$ and responses $\mathbf{u}$ are projected to be reduced inputs $\mathbf{p}\in\mathbb{R}^{n_\text{r}}\times\mathcal{T}$ and outputs $\mathbf{q}\in\mathbb{R}^{n_\text{r}}\times\mathcal{T}$, both of dimensionality of $n_\text{r}$. In practice, the truncated POD basis is highly effective in capturing response signal energy, and it is usually only necessary to keep a small number of POD mode vectors, i.e., $n_\text{r}\ll n_\text{{DoF}}$. This allows the $\mathbf{p}$ and $\mathbf{q}$ to be of very low dimensionality.

Concerning that both the reduced inputs and outputs have potentially a large number of time steps, due to the long time horizon of interest, wavelet transformation is thus introduced as a downsampling approach to reduce the time series \cite{le2015reduced,wang2020knowledge}. This is performed by filtering the time series with a scaling function $\phi(\tau; t)$:
\begin{align}
\label{eq:wave_approx}
    \mathbf{c}_\mathbf{q}(\tau) = \int_t \mathbf{q}(t)\phi(\tau; t) \text{d}t\\
    \mathbf{c}_\mathbf{p}(\tau) = \int_t \mathbf{p}(t)\phi(\tau; t) \text{d}t
\end{align}
$\mathbf{c}_\mathbf{q}(\tau)\in\mathbb{R}^{n_\text{r}\times n_\tau}$ and $\mathbf{c}_\mathbf{p}(\tau)\in\mathbb{R}^{n_\text{r}\times n_\tau}$ are wavelet coefficient series of reduced outputs and inputs, both are reduced to a length of $n_\tau$; $\tau$ is the time shift parameter, which also represents the time index in the wavelet coefficient series. In implementation, it is possible to properly select the scaling function $\phi(\tau; t)$ to significantly reduce the length of time series without noticeable loss in accuracy in reproducing the original signals.

The wavelet coefficient series $\mathbf{c}_\mathbf{q}$ and $\mathbf{c}_\mathbf{p}$ will be used to calibrate the LSTM network. Both the size of $\mathbf{c}_\mathbf{q}$ and $\mathbf{c}_\mathbf{p}$ are being significantly reduced in spatial dimensionality and series length, thus drastically diminishing the computational effort in network calibration and the memory demand on graphics processing unit (GPU). The reduced sequence length also improves LSTM accuracy by mitigating error accumulation during recursive inference. 

\section{Variation LSTM with augmented inputs}\label{Sec:uq}
\subsection{Integrating aleatoric uncertainty as augmented inputs}\label{Sec:Aleatory}
Aleatoric uncertainty under the problem setting of this work includes record-to-record variability, $\textbf{R}$, and system uncertainty, $\mathbf{\Theta}$, as described in Eq.~(\ref{eq1}). The LSTM network allows explicit propagation of the record-to-record variability, carried by the input series $\mathbf{c}_\mathbf{p}$ obtained from stochastic excitation. Monte Carlo simulation, i.e., repeated response simulation based on LSTM inference, directly offers the response samples under the effect of record-to-record variability. The computationally inexpensive LSTM network makes this Monte Carlo simulation process a tractable method for uncertainty propagation. 

To cope with system uncertainty $\mathbf{\Theta}$, its realization $\boldsymbol{\theta}$ is taken as augmented inputs, along with $\mathbf{c}_\mathbf{p}$, to the LSTM network. Unlike $\mathbf{c}_\mathbf{p}$, which is a stochastic process, the $\boldsymbol{\theta}$ includes time-invariant features representing the nonlinear dynamic structural system. From this aspect, $\boldsymbol{\theta}$, if treated as input, should be constant series. To this end, $\boldsymbol{\theta}$ is replicated to be constant series of the same length as $\mathbf{c}_\mathbf{p}$ and concatenated to $\mathbf{c}_\mathbf{p}$, making it to be augmented inputs $\left[\mathbf{c}_\mathbf{p}^\text{T}~~~\boldsymbol{\theta}^\text{T} \right]^\text{T}$, with the superscript T indicating transpose. As such, the LSTM network will be calibrated to capture the system uncertainty along with record-to-record variability. It is worth noting that although there exists an alternative of considering an additional model that maps system uncertainty to neural network parameters, as is mentioned for the NARX models in the introduction, its implementation for deep learning architectures is usually extremely challenging due to the large number of neural network parameters involved. The approach of using augmented inputs, on the other hand, is scalable to machine learning models with arbitrarily large size. In addition, for most cases where typically a few random parameters are involved \cite{chuang2022framework}, this approach offers the most straightforward feature fusion to address the system uncertainty with a slight increase in the input dimensionality. In cases where the system uncertainty is of high-dimensionality, sensitivity analysis along with principal component analysis can be efficiently performed by leveraging the data generated for network calibration, to identify a few mostly critical random parameters. Thereafter, only the most critical random parameters will be added to the augmented inputs. This captures the full system uncertainty while avoiding the high computational and memory demands of increased input dimensionality.

\subsection{Simulate epistemic uncertainty through Monte Carlo dropout}\label{Sec:Epistemic}
A standard machine learning approach, once calibrated, yields a deterministic network characterized by its optimized trainable parameters $\hat{\mathbf{w}}$, which produces predictions when provided with input data. However, these predictions do not inherently convey quantitative information regarding data availability or the extent of training effort invested in model calibration. This implies a risk of being overly confident in the obtained training outcome, particularly in cases with insufficient data or training effort. Indeed, predictions with abundant data and sufficient training should have higher confidence, while those suffering from data sparsity or insufficient training will receive lower confidence. Epistemic uncertainty is associated with the concept of confidence within this context, which quantifies the remaining uncertainty within a trained network, typically decreasing as data availability and training efforts increase. 

\subsubsection{Theoretical foundations of variational inference}
By definition, given a training dataset $\mathcal{D}_\text{t}$ containing $n_\text{t}$ realizations $(\mathbf{f}_i, \boldsymbol{\theta}_i,\mathbf{u}_i)$ selected from the entire dataset $\mathcal{D}$, the epistemic uncertainty in trainable parameters can be described as a posterior distribution $p_{\mathbf{W}|\textbf{D}_\text{t}}(\mathbf{w}|\mathcal{D}_\text{t})$. 
Here, $\textbf{D}_\text{t}$ represents $n_\text{t}$ independent and identically distributed (i.i.d.) random triplets ($\textbf{F}$, $\mathbf{\Theta}$, $\textbf{U}$), with $\mathcal{D}_\text{t}$ being a specific realization of $\textbf{D}_\text{t}$. Once the posterior $p_{\mathbf{W}|\textbf{D}_\text{t}}(\mathbf{w}|\mathcal{D}_\text{t})$ is determined, the epistemic uncertainty in the predicted response $\hat{\mathbf{u}}$ given a new excitation $\hat{\textbf{f}}$ and a system realization $\hat{\boldsymbol{\theta}}$ is quantified by the following posterior predictive distribution:
\begin{equation}
\label{eq:epistemic_pred}
    p_{\mathbf{U}|\textbf{F},\mathbf{\Theta},\textbf{D}_\text{t}}(\hat{\mathbf{u}}|\hat{\textbf{f}},\hat{\boldsymbol{\theta}},\mathcal{D}_\text{t})
    =\int_{\mathbf{w}}     
    p_{\mathbf{U}|\mathbf{W},\textbf{F},\mathbf{\Theta}}(\hat{\mathbf{u}}|\mathbf{w},\hat{\textbf{f}},\hat{\boldsymbol{\theta}})
   p_{\mathbf{W}|\textbf{D}_\text{t}}(\mathbf{w}|\mathcal{D}_\text{t}) \text{d}\mathbf{w}
\end{equation}
where $p_{\mathbf{U}|\mathbf{W},\textbf{F},\mathbf{\Theta}}$ is the distribution of the predicted response $\hat{\mathbf{u}}$ by the model (i.e., the LSTM network). Since the LSTM network provides a deterministic one-to-one mapping, this distribution is represented by a Dirac delta function centered at $\hat{\mathbf{u}}$. It is seen the posterior $p_{\mathbf{W}|\textbf{D}_\text{t}}(\mathbf{w}|\mathcal{D}_\text{t})$ is the key to this probabilistic prediction, which, once obtained, Eq.~(\ref{eq:epistemic_pred}) can be efficiently evaluated, e.g., by Monte Carlo simulation. However, unlike simple conjugate Bayesian models where the posterior admits an analytical form, in neural networks, the complex architecture and nonlinear dependence of the model output on the trainable parameters render the likelihood non-conjugate to the prior. Consequently, the posterior distribution cannot be obtained in closed form. This makes direct posterior inference intractable. 

To this end, variational inference is employed. The core idea is to approximate the posterior through a more tractable variational distribution $q_\mathbf{W}(\mathbf{w})$ by minimizing their discrepancy, which is typically defined as the Kullback–Leibler divergence (KL-divergence): 
\begin{equation}
\label{eq:kl}
    D_{KL}[q_\mathbf{W}(\mathbf{w})||p_{\mathbf{W}|\textbf{D}_\text{t}}(\mathbf{w}|\mathcal{D}_\text{t})]=\mathbb{E}_{q_\mathbf{W}(\mathbf{w})}\log \frac{q_\mathbf{W}(\mathbf{w})}{p_{\mathbf{W}|\textbf{D}_\text{t}}(\mathbf{w}|\mathcal{D}_\text{t})}
\end{equation}
For implementation purposes, $p_{\mathbf{W}|\textbf{D}_\text{t}}(\mathbf{w}|\mathcal{D}_\text{t})$ in Eq.~(\ref{eq:kl}) is replaced by considering the Bayes' theorem:
\begin{equation}
\label{eq:bs}
p_{\mathbf{W}|\textbf{D}_\text{t}}(\mathbf{w}|\mathcal{D}_\text{t})=\frac{p_{\textbf{D}_\text{t}|\mathbf{W}}(\mathcal{D}_\text{t}|\mathbf{w})p_{\mathbf{W}}(\mathbf{w})}{p_{\textbf{D}_\text{t}}(\mathcal{D}_\text{t})}
\end{equation}
where $p_{\textbf{D}_\text{t}|\mathbf{W}}(\mathcal{D}_\text{t}|\mathbf{w})$ is known as the likelihood function; $p_{\mathbf{W}}(\mathbf{w}$) is the prior distribution, representing the statistics of the trainable parameters before the network has gained any information from the dataset; $p_{\textbf{D}_\text{t}}(\mathcal{D}_\text{t})$ is the joint distribution of the training data. Substitute Eq.~(\ref{eq:bs}), the Bayes' theorem, into Eq.~(\ref{eq:kl}), the KL-divergence, one obtains:
\begin{align}
D_{KL}[q_\mathbf{W}(\mathbf{w})||p_{\mathbf{W}|\textbf{D}_\text{t}}(\mathbf{w}|\mathcal{D}_\text{t})]
&= \mathbb{E}_{q_{\mathbf{W}}(\mathbf{w})}\log q_{\mathbf{W}}(\mathbf{w})
 - \mathbb{E}_{q_{\mathbf{W}}(\mathbf{w})}\log
 \frac{p_{\mathbf{D}_\text{t}|\mathbf{W}}(\mathcal{D}_\text{t}\mid\mathbf{w})p_{\mathbf{W}}(\mathbf{w})}
 {p_{\mathbf{D}_\text{t}}(\mathcal{D}_\text{t})} \\
&= \log p_{\mathbf{D}_\text{t}}(\mathcal{D}_\text{t})
 - \mathbb{E}_{q_{\mathbf{W}}(\mathbf{w})}\log p_{\mathbf{D}_\text{t}|\mathbf{W}}(\mathcal{D}_\text{t}\mid\mathbf{w})
 + \mathbb{E}_{q_{\mathbf{W}}(\mathbf{w})}\log
 \frac{q_{\mathbf{W}}(\mathbf{w})}{p_{\mathbf{W}}(\mathbf{w})} \\
&= \log p_{\mathbf{D}_\text{t}}(\mathcal{D}_\text{t})
 - \left(
 \mathbb{E}_{q_{\mathbf{W}}(\mathbf{w})}\log p_{\mathbf{D}_\text{t}|\mathbf{W}}(\mathcal{D}_\text{t}\mid\mathbf{w})
 - D_{KL}\left[q_{\mathbf{W}}(\mathbf{w})||p_{\mathbf{W}}(\mathbf{w})\right]
 \right)
\end{align}
It is observed that the first term $\log p_{\mathbf{D}_\text{t}}(\mathcal{D}_\text{t})$ is a constant, thus can be removed without changing the solution to the optimization problem. It is seen that minimizing $D_{KL}[q_\mathbf{W}(\mathbf{w})||p_{\mathbf{W}|\textbf{D}_\text{t}}(\mathbf{w}|\mathcal{D}_\text{t})]$ is equivalent to maximizing the remaining terms in the equation, known as the log Evidence Lower Bound:
\begin{equation}
\label{eq:elbo}
\mathcal{L}(\mathbf{w}) = 
 \mathbb{E}_{q_{\mathbf{W}}(\mathbf{w})}\log p_{\mathbf{D}_\text{t}|\mathbf{W}}(\mathcal{D}_\text{t}\mid\mathbf{w})
 - D_{KL}\left[q_{\mathbf{W}}(\mathbf{w})||p_{\mathbf{W}}(\mathbf{w})\right]
\end{equation}

\subsubsection{Variational inference through Monte Carlo dropout}
Gal and Ghahramani \cite{gal2015dropout} propose that the variational distribution $q_\mathbf{W}(\mathbf{w})$ can be conveniently realized through Monte Carlo dropout. This is implemented simply by imposing a binary mask for rows of weight matrices of layers where dropout is to be implemented:
\begin{equation}
    \hat{\mathbf{w}}_{j,k}=z_k\mathbf{w}_{j,k}
\end{equation}
where $\mathbf{w}_{j,k}$ is $k$th row of the $j$th matrix $\mathbf{w}_j$ collected in the trainable parameters $\mathbf{w}$; 
$z_k$ is a Bernoulli random variable with probability $P(z_k = 0) = p_j$, where $p_j$ denotes the dropout rate, taking as identical for all $k$. 
As such, each row of the weight matrix $\mathbf{w}_j$ has a probability of $p_{j}$ being masked out, equivalent to dropping the corresponding input from the previous layer. So far, this dropout mechanism has formulated a probabilistic model for the weight matrices, which, although easy to implement, introduces discrete random variables that are not straightforward for further derivation. To facilitate further derivation, it is convenient to smooth the probabilistic model of weight matrices by Gaussian kernels, making it a Gaussian mixture distribution \cite{gal2015dropout} with a small bandwidth $\sigma_\mathbf{w}$:
\begin{equation}
    q(\mathbf{w}_{j,k}) = (1-p_{j})\mathcal{N}(\mathbf{w}_{j,k},\sigma_\mathbf{w}^2\mathbf{I}_{\mathbf{w}_{j,k}})+p_{j}\mathcal{N}(\mathbf{0},\sigma_\mathbf{w}^2\mathbf{I}_{\mathbf{w}_{j,k}})
\end{equation}
where $\mathcal{N}(\cdot)$ indicates normal distribution; $\mathbf{I}_{\mathbf{w}_{j,k}}$ is an identity matrix whose dimension equals the length of $\mathbf{w}_{j,k}$. Next, without loss of generality, the prior distribution $p(\mathbf{w}_{j,k})$ can be taken as a Gaussian distribution with zero mean and the covariance matrix to be the identity. As such, it can be shown that the second term in Eq.~(\ref{eq:elbo}) can be simplified to be \cite{gal2015dropout}:
\begin{align}
\label{eq:kl_2_norm}
    D_{KL}\left[q_{\mathbf{W}}(\mathbf{w})||p_{\mathbf{W}}(\mathbf{w})\right]=\sum_{j} \frac{1-p_{j}}{2} ||\mathbf{w}_{j}||_2^2+C(\sigma_\mathbf{w})
\end{align}
where $C(\sigma_\mathbf{w})$ is a constant.

The log likelihood $\log  p_{\textbf{D}_\text{t}|\mathbf{W}}(\mathcal{D}_\text{t}|\mathbf{w})$ in Eq.~(\ref{eq:elbo}) in principle can be derived by $p_{\mathbf{U}|\mathbf{W},\textbf{F},\mathbf{\Theta}}$. However, the deterministic nature of the LSTM network makes this distribution a Dirac delta function, which leads to infinite log likelihood values. To facilitate a tractable derivation, we adopt a Gaussian form for $p_{\mathbf{U}|\mathbf{W},\textbf{F},\mathbf{\Theta}}$, with mean given by the neural network output $\mathcal{NN}(\mathbf{w}; \mathbf{f}, \boldsymbol{\theta})$ and variance $\sigma^2$, i.e., additive zero-mean Gaussian noise with standard deviation $\sigma$ is considered. After the derivation, we can let $\sigma^2\rightarrow 0$ to recover the Dirac delta form of $p_{\mathbf{U}|\mathbf{W},\textbf{F},\mathbf{\Theta}}$ to reconcile with the deterministic nature of the LSTM network. The Gaussian likelihood reads:
\begin{equation}
\label{eq:nn_pred_normal}
    p_{\mathbf{U}|\mathbf{W},\textbf{F},\mathbf{\Theta}}(\hat{\mathbf{u}}|\mathbf{w},\hat{\textbf{f}},\hat{\boldsymbol{\theta}})
    =\mathcal{N}(\mathcal{NN}(\mathbf{w};\hat{\textbf{f}},\hat{\boldsymbol{\theta}}),\sigma^2\mathbf{I}_{\hat{\mathbf{u}}})
\end{equation}
$\mathbf{I}_{\hat{\mathbf{u}}}$ is an identity matrix with a dimension of $n_rn_\tau$, i.e., the total number of entries contained in $\hat{\mathbf{u}}$; The log likelihood $\log p_{\textbf{D}_\text{t}|\mathbf{W}}(\mathcal{D}_\text{t}|\mathbf{w})$ can be directly obtained by substituting each datum into Eq.~(\ref{eq:nn_pred_normal}) and noting that the joint likelihood is the product of the likelihood of each datum, by reasonably assuming that each datum is independent:
\begin{align}
\log p_{\textbf{D}_\text{t}|\mathbf{W}}(\mathcal{D}_\text{t}|\mathbf{w})&=\sum_{i=1}^{n_\text{t}} \log p_{\mathbf{U}|\mathbf{W},\textbf{F},\mathbf{\Theta}}(\mathbf{u}_i|\mathbf{w},\textbf{f}_i,\boldsymbol{\theta}_i)\\
\label{eq:log_like_normal}
&=-\frac{n_\text{t}n_rn_\tau}{2}\log(2\pi)-\frac{n_\text{t}n_rn_\tau}{2}\log\sigma^2-\sum_{i=1}^{n_\text{t}}\frac{1}{2\sigma^2}||\mathbf{u}_i-\mathcal{NN}(\mathbf{w};\textbf{f}_i,\boldsymbol{\theta}_i)||_2^2
\end{align}
Substitute Eqs.~(\ref{eq:kl_2_norm}) and~(\ref{eq:log_like_normal}) into Eq.~(\ref{eq:elbo}), ignore constant terms that are not changing with trainable parameters $\mathbf{w}$, and normalize by $n_\text{t}/\sigma^2$, one obtains the log evidence lower bound under the dropout approximation:
\begin{equation}
\label{eq:elbo_dropout}
\mathcal{L}(\mathbf{w}) = -\mathbb{E}_{q_{\mathbf{W}}(\mathbf{w})}\frac{1}{2n_\text{t}}\sum_{i=1}^{n_\text{t}}||\mathbf{u}_i-\mathcal{NN}(\mathbf{w};\textbf{f}_i,\boldsymbol{\theta}_i)||_2^2-\sum_{j} \frac{\sigma^2 (1-p_{j})}{2n_\text{t}} ||\mathbf{w}_{j}||_2^2
\end{equation}
It is interesting to note how this approximated log evidence lower bound resembles the regular square loss with dropout and L2 regularization, except for a negative sign ensuring Eq.~(\ref{eq:elbo_dropout}) is a maximization problem that is consistent with Eq.~(\ref{eq:elbo}). This makes training a variational LSTM network the same as training a regular LSTM network with dropout and considering the square loss with regularization. The variance of the additional Gaussian noise, $\sigma^2$ in Eq.~(\ref{eq:nn_pred_normal}), controls the regularization term, and letting $\sigma^2\rightarrow0$, the regularization term vanishes. The developed framework allows one to choose a non-zero $\sigma^2$ if the training data contain noise, e.g., from experiments or sensors. In this work, we let $\sigma^2\rightarrow0$ to maintain consistency with the deterministic mapping of the LSTM architecture. Although getting rid of this regularization term might typically increase the risk of overfitting, it should be considered that the dropout is also, in nature, a regularization technique preventing overfitting \cite{srivastava2014dropout}, making the additional L2 regularization may not often be necessary. 

As such, in implementation, one first performs dropout to obtain network realizations. Next, the objective function Eq.~(\ref{eq:elbo_dropout}) and its derivatives are evaluated. The trainable parameters $\mathbf{w}$ are subsequently updated. This parameter-updating cycle is performed iteratively until a satisfactory result is reached. To alleviate the computational burden of full-batch gradient evaluation and to introduce stochasticity that helps escape suboptimal local minima, the optimization is typically performed in a mini-batch manner, whereby gradients are estimated from randomly sampled subsets of the training data at each iteration. Lastly, it should be noted that when dropout is applied to recurrent neural networks, e.g., LSTM in this paper, the same parameter realization must be considered for all the time steps \cite{gal2016theoretically} to avoid very weak signals due to a great number of repeated dropouts \cite{gal2015dropout}. Once calibrated, the epistemic uncertainty can be simulated by simply sampling network realizations with the Monte Carlo dropout.



\begin{figure}
    \centering
    \includegraphics[width=0.7\linewidth]{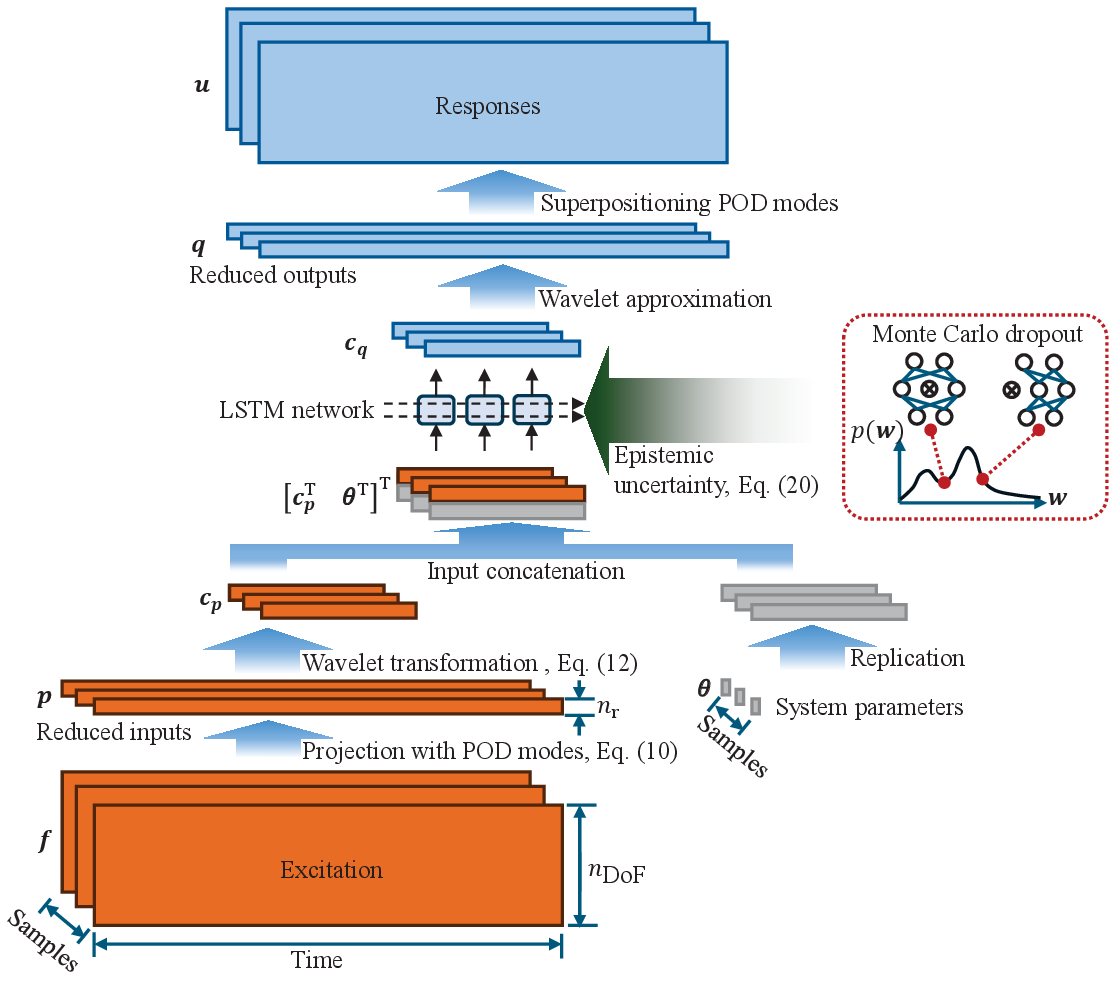}
    \caption{The developed framework}
    \label{fig:framework}
\end{figure}

\section{Summary of the metamodeling approach}
The developed metamodeling approach leverages the LSTM network with augmented inputs to propagate aleatoric uncertainty, including record-to-record variability and system uncertainty, and uses Monte Carlo dropout to simulate epistemic uncertainty. To manage high dimensionality and extended durations, POD-based dimensionality reduction and wavelet-based downsampling are used to reduce the size of the excitation and responses. The process of the developed metamodeling approach is shown in Figure~\ref{fig:framework} and summarized below for implementation:
\begin{enumerate}
    \item Data generation: Generate $n$ realizations of stochastic excitation and structural systems $\mathbf{f}_i, \boldsymbol{\theta}_i, i=1,2,...,n$, and perform high-fidelity simulation by numerically solving the dynamic equilibrium, Eq.~(\ref{eq1}), to obtain their response samples, $\textbf{u}_i, i=1,2,...,n$. This gives a dataset $\mathcal{D} = \{\textbf{f}_i, \boldsymbol{\theta}_i, \textbf{u}_i: i=1,2,...,n\}$. Select randomly from the dataset $\mathcal{D}$ for the calibration set $\mathcal{D}_c$ to calibrate the metamodel, and leave the rest of the data for the testing set $\mathcal{D}_e$ to evaluate the generalizability of the calibrated metamodel to unseen data. In the calibration set $\mathcal{D}_c$, randomly select a small group of data as a validation set $\mathcal{D}_v$ to monitor potential overfitting issues, and leave the rest of the data for the training set $\mathcal{D}_t$ for gradient evaluation.
    \item Data processing: Normalization statistics (e.g., mean, maximum/minimum) are first evaluated from the training set $\mathcal{D}_t$. Next, POD basis $\boldsymbol{\phi}$ is extracted from response snapshots selected randomly or uniformly in time from the response samples in $\mathcal{D}_t$. This is done by firstly obtaining all POD mode vectors by singular value decomposition for the matrix collecting snapshots, and then truncating to keep only significant POD mode vectors, i.e., those corresponding to the first $n_r$ largest singular values. Projections by Eqs.~(\ref{pod_u})~and~(\ref{pod_f}) are performed for the excitations and responses for all the data $\mathcal{D}$ to obtain reduced inputs and outputs,  $\{\textbf{p}_i, \textbf{q}_i: i=1,2,...,n\}$. Next, wavelet-based downsampling with a properly selected scaling function is leveraged to further reduce the length of these time series. This gives the wavelet coefficient series with reduced length, $\{\mathbf{c}_{\mathbf{p},i}, \mathbf{c}_{\mathbf{q},i}: i=1,2,...,n\}$. Finally, the parameters describing realizations of system uncertainty, $\boldsymbol{\theta}_i$, are paired with the wavelet coefficients of reduced inputs, replicated to match the length of the coefficient series, and concatenated to be the augmented inputs $\left[\mathbf{c}_{\mathbf{p},i}^\text{T}~~~\boldsymbol{\theta}_i^\text{T} \right]^\text{T}$. This forms the data format to be learned by the LSTM network: $\{\left[\mathbf{c}_{\mathbf{p},i}^\text{T}~~~\boldsymbol{\theta}_i^\text{T} \right]^\text{T}, \mathbf{c}_{\mathbf{q},i}: i=1,2,...,n\}$.
    \item Network calibration: LSTM is calibrated to a one-to-one mapping $\left[\mathbf{c}_{\mathbf{p},i}^\text{T}~~~\boldsymbol{\theta}_i^\text{T} \right]^\text{T}\xrightarrow{}\mathbf{c}_{\mathbf{q},i}$, with Monte Carlo dropout incorporated to simulate epistemic uncertainty. Stochastic gradient descent and its variants can be leveraged to train the LSTM network by optimizing the loss function in Eq.~(\ref{eq:elbo_dropout}), which is equivalent to maximizing the log evidence lower bound in Eq.~(\ref{eq:elbo}). During the training, the loss function will also be evaluated based on the validation set $\mathcal{D}_v$ to monitor the generalizability performance of the LSTM network, triggering early stopping if unacceptable overfitting is observed. This iterative training process is allowed to continue until a satisfactory LSTM network is obtained.
    \item Uncertainty propagation: To perform simulation for any excitation samples and system realizations, e.g., those in the testing set $\mathcal{D}_t$, the same data processing described in Step 2 will be performed. This gives augmented inputs carrying aleatoric uncertainty. Simultaneously, the Monte Carlo dropout will be performed to generate multiple realizations of the calibrated LSTM network. Each of the augmented inputs will be fed into these realizations of the calibrated LSTM network to obtain response samples representing epistemic uncertainty. This, overall, allows efficient propagation of aleatoric and epistemic uncertainty to dynamic responses.
\end{enumerate}

\section{Case study}
In this section, we demonstrate the applicability of the developed metamodeling framework with a single DoF (SDOF) Bouc-Wen system under seismic excitation, a shear building model subjected to seismic excitation, and a 2D steel framework subjected to wind excitation. All the numerical computations involved in this paper were performed on a Personal Computer equipped with an Intel(R) Core(TM) i9-14900K CPU (3.20 GHz) and 64.0 GB of RAM. The LSTM training was executed on a single NVIDIA GeForce RTX 4090 GPU with 24.0 GB of VRAM.
\subsection{Case 1: SDOF Bouc-Wen system subjected to stochastic seismic excitation}
\subsubsection{Structural system}
\begin{figure}
    \centering
    \includegraphics[width=0.7\linewidth]{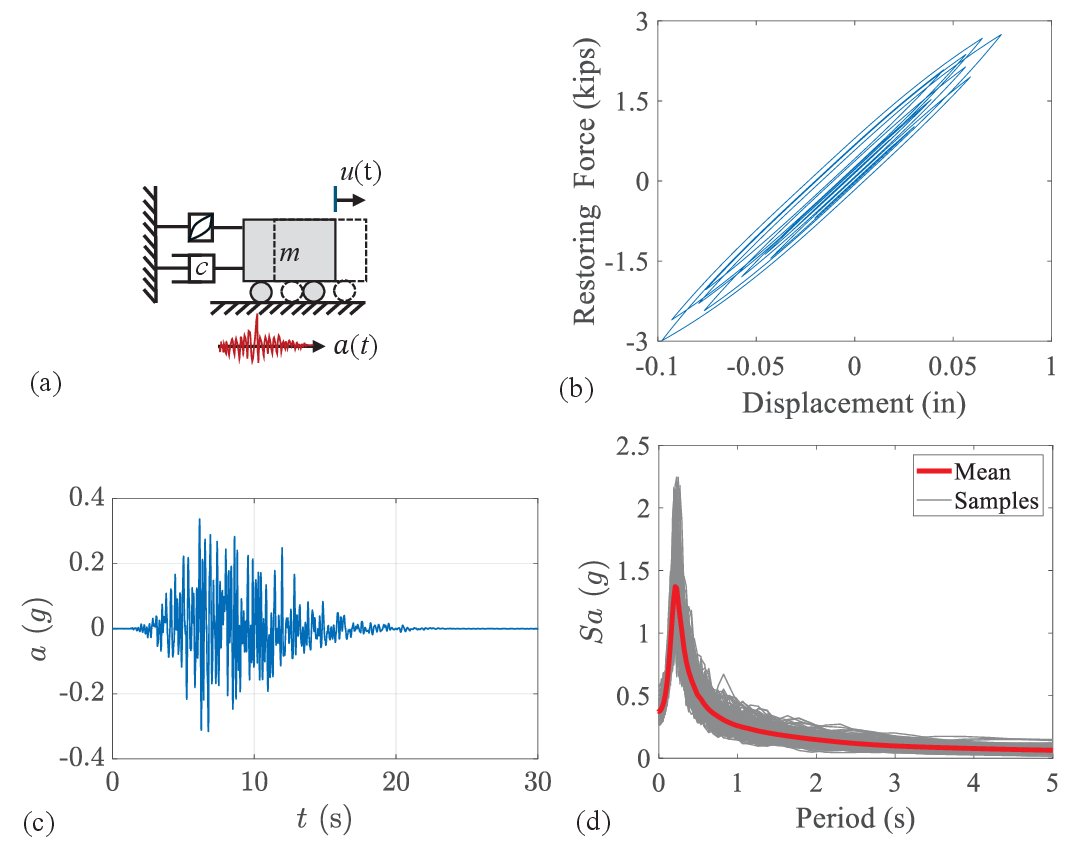}
    \caption{(a) The SDOF Bouc-Wen system subjected to ground motion in Case 1; (b) A realization of the hysteresis curve; (c) A realization of ground motion; (d) Mean spectrum and spectra samples.}
    \label{fig:Case1Structure}
\end{figure}

This case study considers an SDOF Bouc-Wen system with parametric uncertainty, subjected to stochastic seismic excitation, as shown in Figure~\ref{fig:Case1Structure} (a). The Bouc-Wen system is widely recognized for its versatility in representing the non-linear hysteretic behavior seen in materials, components, and structures. The deterministic form of the equation of motion for the considered SDOF Bouc-Wen system is given by:
\begin{equation}
    \ddot{u}(t) + 2\zeta\omega\dot{u}(t) + \omega^2\left[\rho u(t) + (1 - \rho)z(t)\right] = -a(t), 
\end{equation}
where $a(t)$ is the ground motion acceleration; $u(t)$, $\dot{u}(t)$, $\ddot{u}(t)$ are the displacement response and its time derivatives, i.e., velocity, and acceleration; $\zeta$ and $\omega$ are respectively the damping ratio and circular frequency in the linear elastic sense; $\rho$ is the post-yield stiffness ratio; $z(t)$ is the hysteretic displacement, which is governed by the equation:
\begin{equation}
    \dot{z}(t) = \gamma\dot{u}(t) - \alpha|\dot{u}(t)||z(t)|^{n-1}z(t) - \beta\dot{u}(t)|z(t)|^n 
\end{equation}
where $\dot{z}(t)$ is the first time derivative of the hysteretic displacement $z(t)$; parameters $\alpha$, $\beta$, $\gamma$, and $n$ control the hysteretic behaviors of the system. In this case study, the fundamental frequency $\omega$ and the hysteretic parameter $\alpha$ are treated as random variables following uniform distributions, and the lower and upper bounds as defined in Table \ref{tab:Case1Parameters}. The remaining parameters are held constant at $\zeta = 0.02$, $\rho = 0$, $n = 2$, $ \gamma = 1$ and $\beta = 0$. Figure~\ref{fig:Case1Structure} (b) displays a realization of the hysteresis curve.

\begin{table}[]
\caption{Random Parameters in Bouc-Wen System}\label{tab:Case1Parameters}
\begin{tabular*}{\tblwidth}{@{}LLLL@{}}
\toprule
   Parameters&  Distribution& Mean &Support \\ 
\midrule
 $\omega$&  Uniform& 5.97& [5.373, 6.567] \\
 $\alpha$& Uniform& 50&[45, 55] \\
 
\bottomrule
\end{tabular*}
\end{table}

The excitation was generated by the stochastic ground motion model from \cite{rezaeian2010}. This model generates stochastic ground motion by first passing a white noise series through an SDOF filter with time-varying circular frequency and damping ratio, followed by the application of a time-modulating function to define the evolving amplitude. A high-pass filter is finally applied to remove the unrealistic displacement or velocity residuals to avoid overestimation in long-period components of responses. The parameters of the ground motion model are $I_a = 0.109$ s$\cdot$g, $D_{5-95} = 7.96$ s, $t_\text{mid} = 7.78$ s, $\omega_\text{mid} = 4.66 \times 2 \times \pi$ rad/s, $\omega' = -0.09 \times 2 \times \pi$ rad/s$^2$, and  $\zeta_f = 0.24$, adopted from \cite{rezaeian2010} for the recorded motion at the 1994 Northridge earthquake at the LA 00 station. All notations are kept consistent with the cited reference. Figure~\ref{fig:Case1Structure} (c) and (d) respectively show a ground motion realization and the spectra of ground motions generated by the model.

\subsubsection{Data generation}
\label{sec:case1_data}
The Bouc-Wen equation was modeled and solved in MATLAB R2024b environment. In this case, we consider generating a dataset $\mathcal{D}$ with 200 excitation-system-response samples. To this end, we first generate 200 ground motion realizations based on the aforementioned stochastic ground motion model to represent record-to-record variability. Next, 200 system realizations are generated by sampling the circular frequency $\omega$ and the hysteretic parameter $\alpha$, in modeling the system uncertainty. These 200 system realizations are subsequently paired with the ground motion realizations, and the resulting response time histories were computed using the fourth-order Runge-Kutta algorithm with an error tolerance of $10^{-5}$. 

\subsubsection{Training and testing configuration}
\begin{figure}
    \centering
    \includegraphics[width=0.5\linewidth]{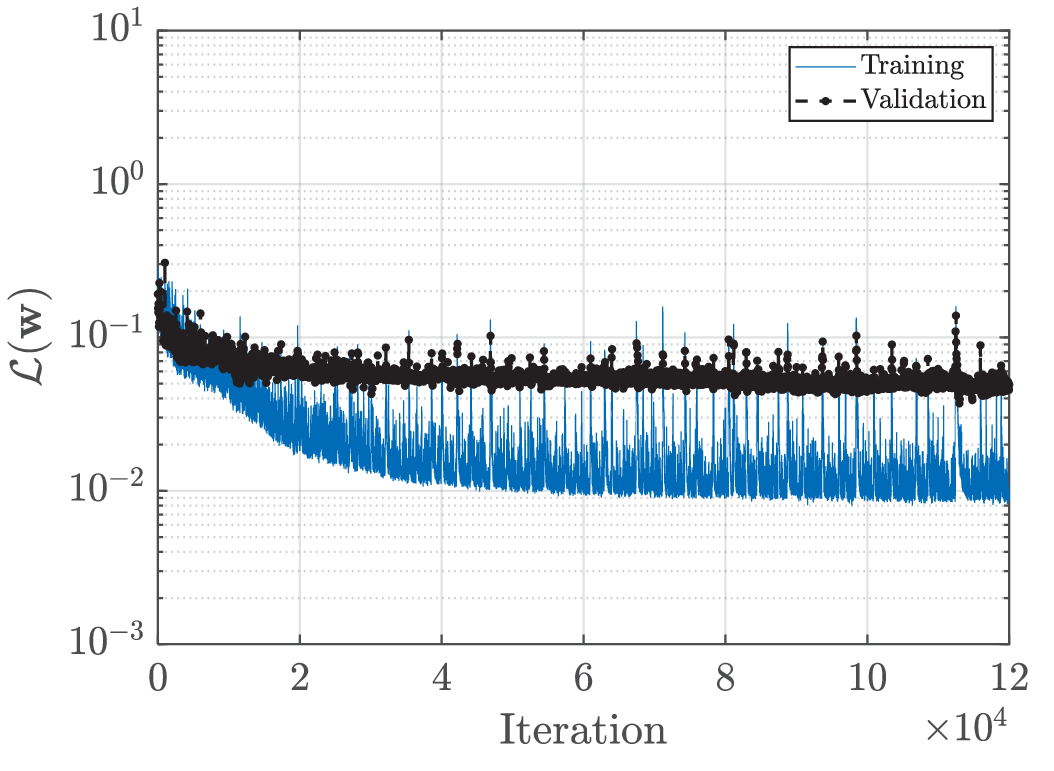}
    \caption{Training and validation loss of metamodel in Case 1}
    \label{fig:caseItraining}
\end{figure}

The metamodel in this case adopts an LSTM architecture, consisting of a sequence input layer, an LSTM layer with 200 hidden units, followed by a fully connected layer. The network comprises three input features, including a ground motion record, concatenated with the two uncertain parameters $\omega$ and $\alpha$ replicated to have the same length as the ground motion record. The output layer gives a single feature, displacement response time history, as the output data.  
The dropout is implemented with a dropout rate of 20 \% for the fully connected layer to introduce uncertainty in modeling epistemic uncertainty to the metamodel  \cite{gal2015dropout}. Next, to calibrate the metamodel, the 200 data were divided by allocating 150 data to the training set, 10 data to the validation set, and 40 data to the testing set.
The training was executed by leveraging the Adam optimizer with a learning rate of 0.002. Loss gradient in each iteration was evaluated based on the training set by considering a mini-batch size of 50, making the training with 3 iterations per epoch. Figure~\ref{fig:caseItraining} shows the evolution of the training and validation loss during training. The training was concluded in a total of 40,000 epochs ($12 \times 10^4$ iterations) after converged training and validation loss were observed. The entire training process for Case 1 took around 671 minutes and 36 seconds. 

\subsubsection{Performance on the testing set}
\begin{figure}
    \centering
    \includegraphics[width=0.7\linewidth]
{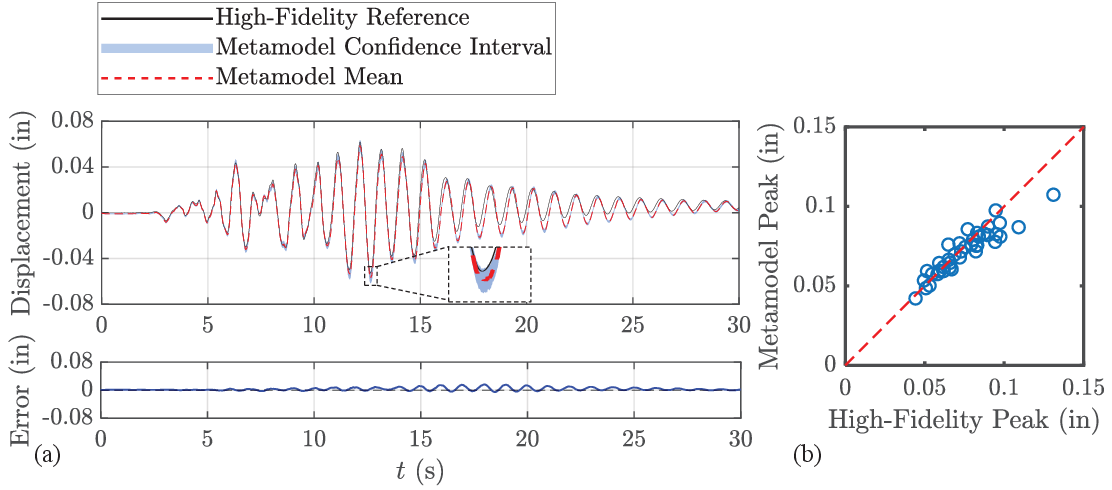}
    \caption{(a) Comparison of the mean and 95\% confidence interval of the displacement response by the metamodel against the high-fidelity reference; (b) Error time history of the mean response by the metamodel; (c) Comparison between peak displacements by the metamodel and by the high-fidelity reference.}
    \label{fig:caseIResults}
\end{figure}
The performance of the metamodel was evaluated on the testing set with 40 samples, with results shown in Figure \ref{fig:caseIResults}. In particular, in Figure \ref{fig:caseIResults} (a), we show the response time history from the metamodel, with a median level of mean square error over the testing set, compared against that from the numerical solver (fourth-order Runge-Kutta method). It is seen that the response reproduced by the metamodel matches well with the numerical solver over the entire time history. Figure \ref{fig:caseIResults} (b) shows the error time history between the response by the metamodel and the numerical solver, where only minor discrepancies are observed. Figure \ref{fig:caseIResults} (c) shows the comparison between the peak responses by the metamodel and the numerical solver for the entire testing set. It is seen that the augmented LSTM successfully reproduced the responses from the dynamic system with a full range of aleatoric uncertainty, including record-to-record variability and system uncertainty. In addition, the metamodel can also simulate the epistemic uncertainty using the Monte Carlo dropout. This is demonstrated by the interval of 95\% confidence, as the shaded area shown in Figure \ref{fig:caseIResults} (a), along with the response time histories. This confidence interval is obtained by generating 100 metamodel realizations and passing the ground motion through these realizations. This gives 100 response time history realizations representing the epistemic uncertainty, from which the corresponding confidence interval can be constructed. This confidence interval reflects the remaining epistemic uncertainty in the metamodel predictions, given the amount of data and training effort provided.

\subsection{Case 2: Nonlinear shear building model subjected to stochastic seismic excitation}
\subsubsection{Structural system}
\begin{figure}
    \centering
    \includegraphics[width=0.5\linewidth]{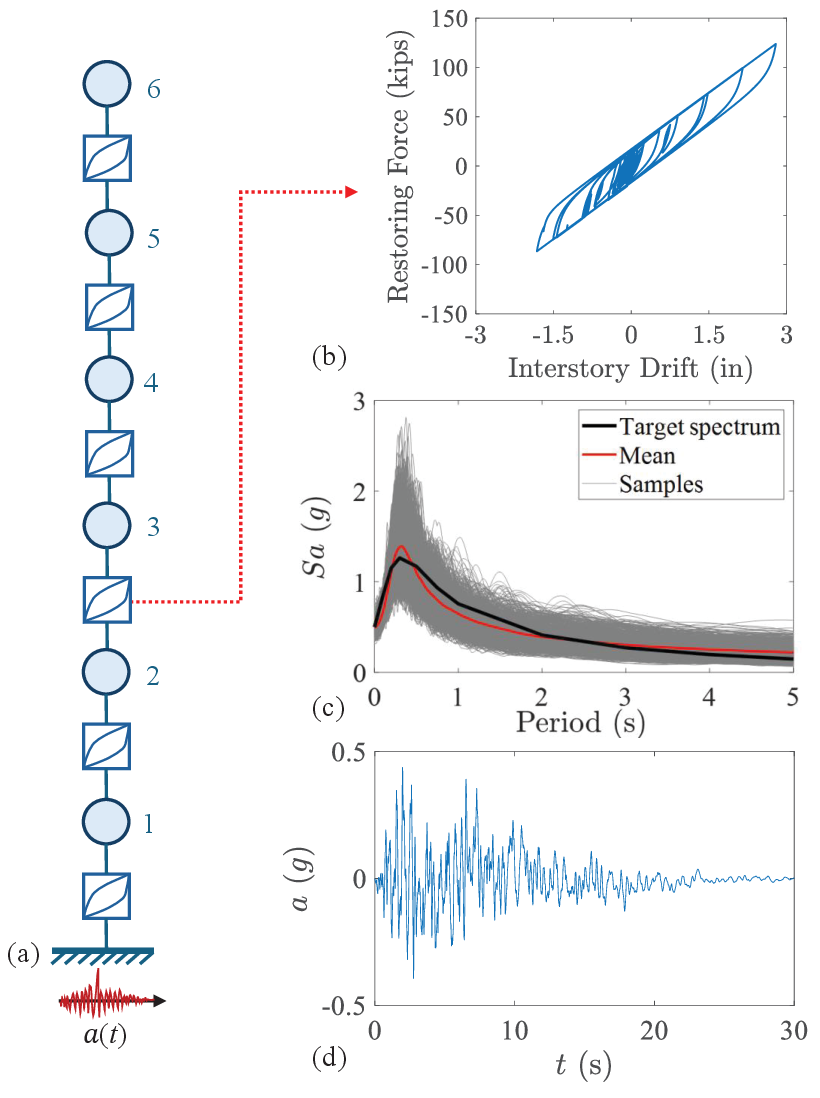}
    \caption{(a) The shear building model in Case 2; (b) A representative inter-story hysteresis curve; (c) The target spectrum, mean and sample spectra of the generated ground motion realizations; (d) A representative ground motion realization.}
    \label{fig:Case2Structure}
\end{figure}

To demonstrate the scalability of the metamodeling approach, a six-story nonlinear shear building model under seismic excitation is considered \cite{McKenna2024quoFEM} as shown in Figure~\ref{fig:Case2Structure} (a). The mass of the model is given by the roof weight $w_r$ and equal floor weights $w$ at each floor. The nonlinear interstory restoration force is modeled through the Giuffre-Menegotto-Pinto model, characterized by the initial stiffness $k_e$, the yield strength $f_\text{y}$, the post-yield stiffness ratio $\rho$, and parameters $R_0$, $cR_1$, and $cR_2$ controlling the transitioning from elastic to plastic branches. This case accounts for the uncertainty of the two most influential parameters: story weight ($w$) and post-yield stiffness ratio ($\alpha$), both of which follow uniform distributions and the lower and upper bounds as defined in Table~\ref{tab:case2Parameters}. The rest parameters take fixed values as initial stiffness $k_e=300$ kip/in (52,538.05 kN/m), the yield strength $f_\text{y}=20$ kip/in$^2$ (137,895.14 kN/m$^2$), $R_0=15$, $cR_1=0.925$, $cR_2=0.15$ , and the roof weight $w_r=60$ kips (266.89 kN). A representative hysteresis curve of this structure is shown in Figure~\ref{fig:Case2Structure} (b), indicating the high-nonlinearity with this case.

\begin{table}[]
\caption{Random Parameters in Nonlinear Shear Building}\label{tab:case2Parameters}
\begin{tabular*}{\tblwidth}{@{}LLLL@{}}
\toprule
   Parameters&  Distribution& Mean &Support \\ 
\midrule
  Story Weight (w) &Uniform& 
     100 kips (444.82 kN) &[50,150] kips ([222.41, 667.23]kN) \\
  Post-Yield Stiffness ratio ($\alpha$) &Uniform& 0.1&[0.05,0.15] \\
 
\bottomrule
\end{tabular*}
\end{table}

\subsubsection{Hazard and ground motion}
\label{sec:case2_gm}
We considered a seismic hazard of 10\% exceedance in 50 years at Loma Preita. Target spectrum was then obtained from USGS (United States Geological Survey) hazard tool \cite{USGS_interactive_hazard} at this hazard level, as shown in Figure~\ref{fig:Case2Structure} (c). We employed the same ground motion generator as mentioned in Case 1 \cite{rezaeian2010}. The parameters for the ground motion generator were taken first from those for the recorded 1989 Loma Preita earthquake reported in \cite{rezaeian2010} and calibrated to the target spectrum as is described in \cite{li2022metamodeling}. This gives the parameters of the ground motion generation model: $I_a = 0.045$ s$\cdot$g, $D_{5-95} = 12.62$ s, $t_m = 4.73$ s, $\omega_\text{mid} = 20.57$ rad/s, $\omega' = -0.08 \times 2 \times \pi$ rad/s$^2$, and $\zeta_f = 0.4801$ representing the Loma Preita earthquake but with the hazard level of 10\% exceedance in 50 years designated. The spectra of all ground motion realizations and their mean are shown in Figure~\ref{fig:Case2Structure} (c) while Figure~\ref{fig:Case2Structure} (d) shows a representative ground motion realization.

\subsubsection{Data generation}
To obtain the dataset needed for metamodel calibration and testing, 1000 ground motion realizations, each with a duration of 30 seconds, were generated through the ground motion model as outlined in Section~\ref{sec:case2_gm}. Next, the 1000 groups of parameters $w$ and $\alpha$ are sampled, compiled to be structural system realizations, and paired with the ground motion realizations. Simulations through direct integration are carried out for the high-fidelity shear building model under these seismic excitations within the OpenSees (the Open System for Earthquake Engineering Simulation) version 3.7.1 \cite{OpenSees2000}. OpenSees Tcl scripts were run in the MATLAB workflow repeatedly for all 1000 ground motion-structural system pairs. The Tcl scripts execute this nonlinear dynamic analysis using the average constant acceleration method. At each time step, the nonlinear equilibrium was solved using an adaptive Newton-Raphson loop; if the standard algorithm failed to converge, the system automatically switched to the modified Newton-Raphson method \cite{McKenna2024quoFEM}. This produces 1000 responses, including the displacement and velocity at each DoF, as well as the restoring force within each story.


\subsubsection{Training and testing configuration}
\begin{figure}
    \centering
    \includegraphics[width=0.5\linewidth]{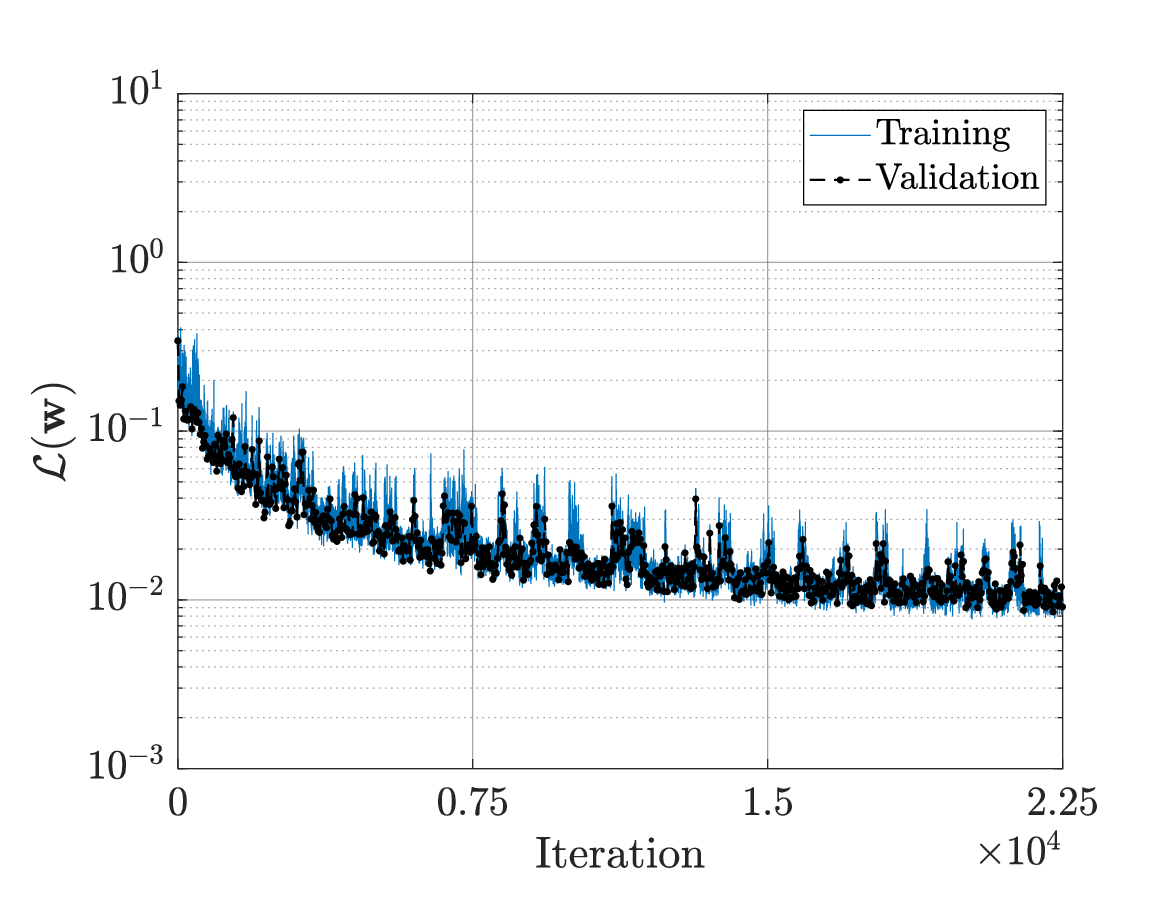}
    \caption{Training and validation loss of metamodel in Case 2}
    \label{fig:Case2Training}
\end{figure}

We considered the same metamodel configuration as in Case 1, where the metamodel is an LSTM network with 200 hidden units. The network input comprises three features: a ground motion and two random parameters. The two random parameters were concatenated to the ground motion realization after being replicated to match its temporal length. The output layer produces the displacement response time histories. To calibrate the LSTM network, we divided the 1000 data into 750 for training, 50 for validation, and the remaining 200 for testing to evaluate the performance of the metamodel on unseen realizations. To simulate the epistemic uncertainty, a dropout rate of 20\% for the fully connected layer was incorporated following the Monte Carlo dropout framework. We considered a mini-batch size of 50, resulting in 15 iterations per epoch. The network was then trained for 1500 epochs ($2.25 \times 10^4$ iterations) with a learning rate of 0.002. The curve of training and validation loss in the training process is shown in Figure~\ref{fig:Case2Training}. Both the training and validation losses converge to a low level, with no evidence of overfitting. The total training time for this case is approximately 191 minutes and 10 seconds. 

\subsubsection{Performance on the testing set}
\begin{figure}
    \centering
    \includegraphics[width=0.7\linewidth]{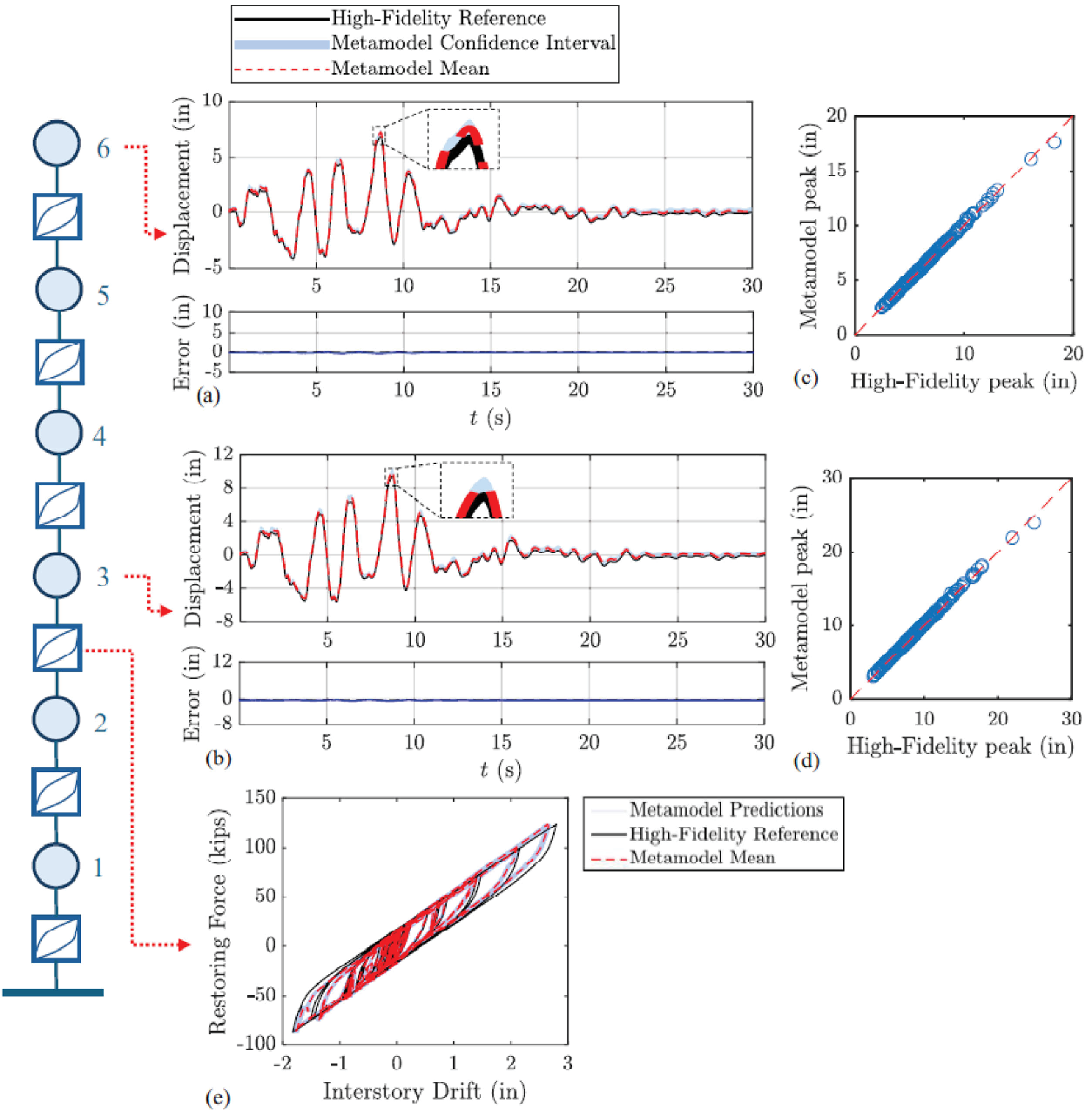}
    \caption{(a)(b) Comparison of displacement response time histories between high-fidelity direct integration and the metamodel, including 95\% confidence intervals (light blue shaded areas), at the 3rd and 6th DoFs; (c)(d) Comparison between the peak displacements by the high-fidelity direct integration and by the metamodel, at the 3rd and 6th DoFs; (e) The hysteresis curves by the high-fidelity direct integration and by the metamodel, as well as 100 realizations of the hysteresis curve under epistemic uncertainty (light blue).}
    \label{fig:Case2Results}
\end{figure}

Figure \ref{fig:Case2Results} shows the displacement responses, including the time histories with median level error and the peak displacements for the entire testing set (plotted with the line of identity) at the 3rd and 6th DoF, as well as the hysteresis curve at the 3rd floor. The hysteresis in Figure~\ref{fig:Case2Results} (e) shows this is a highly nonlinear case. Overall, the metamodel exhibits impressive accuracy after the calibration. It is seen from Figure~\ref{fig:Case2Results} (a) and (b) that the response time histories predicted by the metamodel closely align with the reference solutions obtained via direct integration in OpenSees. The high accuracy is observed for the entire testing set, as can be seen in the comparison of the peak responses by the metamodel and the high-fidelity direct integration shown in Figure~\ref{fig:Case2Results} (c) and (d), where sample points tightly cluster along the line of identity. On the other hand, the intervals of 95\% confidence shown in Figure~\ref{fig:Case2Results} (a) and (b), simulated by the Monte Carlo dropout, reflect the model's predictive certainty under the finite size of the training dataset and the optimization process. In addition, the metamodel allows reproducing the interstory hysteresis curve, as shown in Figure~\ref{fig:Case2Results} (e), which matches perfectly with that provided by the high-fidelity simulation. The ensemble of hysteresis realizations generated by the Monte Carlo dropout (shown in light blue) characterizes the epistemic uncertainty in the predicted nonlinear behavior. This confirms that the metamodeling framework allows for the estimation of the epistemic uncertainty of a full range of responses from displacements to hysteresis.   

\subsection{Case 3: 37-story moment-resisting frame under stochastic wind excitation}
\subsubsection{Structural system}
\begin{figure}
    \centering
    \includegraphics[width=0.7\linewidth]{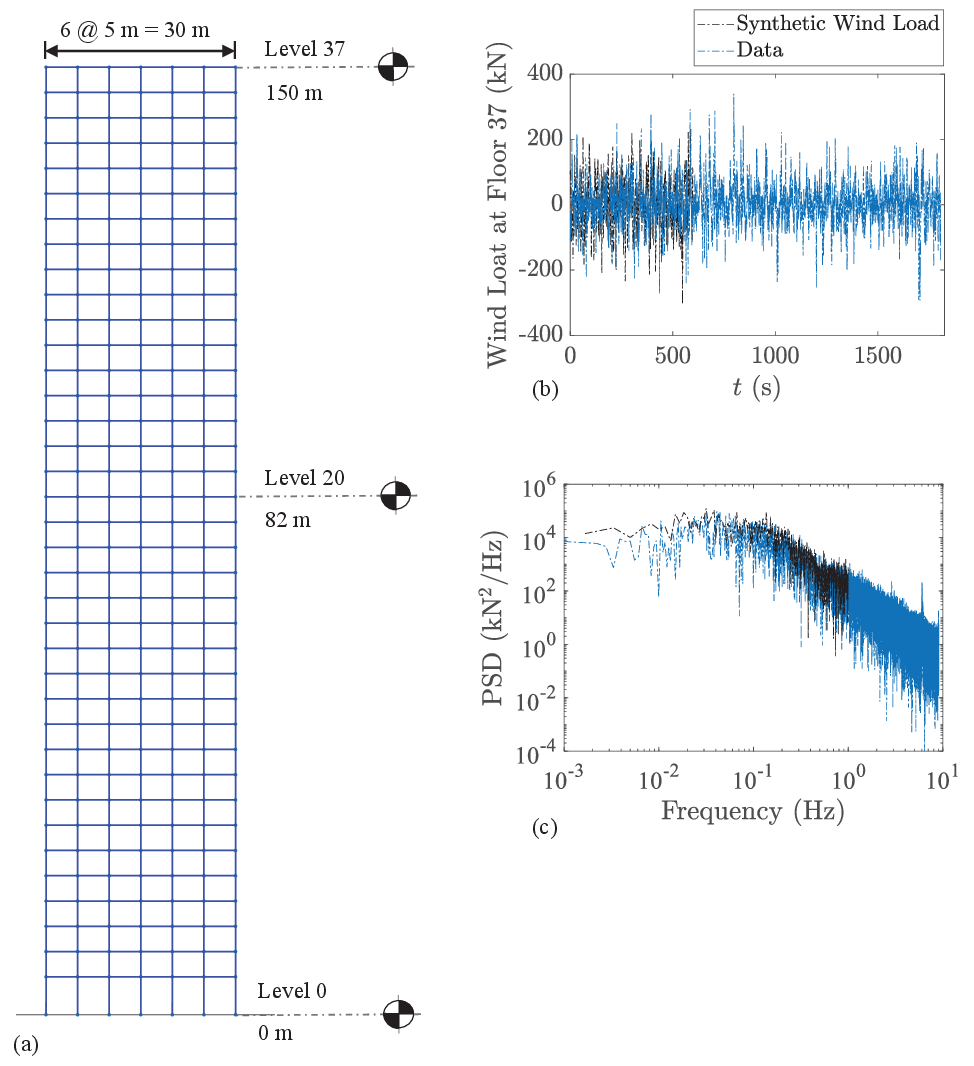}
    \caption{(a) The 37-Story moment-resisting frame considered in Case 3; Comparison between the (b) time histories and (c) power spectral densities (PSD) of synthetic wind load and the wind tunnel data at the top floor.}
    \label{fig:case3Structure}
\end{figure}

In this section, a 37-Story moment resisting frame is subjected to wind excitation to illustrate the scalability of this metamodeling approach to cases with high-dimensional nonlinear structural systems and stochastic excitation. This structure is adopted from \cite{li2024time} and is shown in Figure \ref{fig:case3Structure} (a). The total height of this 2D steel frame is 150 m. The first floor is 6 m in height, and the rest of the floors have a floor height of 4 meters each. Each floor has six spans with an equal width of 5 m. The structural system consists of columns with box sections and AISC standard wide-flange W24 section beams. The dimensions of the beam and column vary with the height of the floor, as summarized in Table~\ref{tab:case3Structure}. The sections are composed of steel material with the yield stress $\sigma_y = 355$ MPa, and Young's modulus $E_s = 200$ GPa. A volumetric mass density of $100$ kg/m$^3$ is considered to account for the superimposed dead loads in addition to the self-weight of the structural members. The inherent damping of the structure was considered as random, as the source of system uncertainty.

\begin{table}[]
\caption{Dimensions of the structural members in Case 3}\label{tab:case3Structure}
\begin{tabular*}{\tblwidth}{@{}LLL@{}}
\toprule
  \textbf{Floor}&  \textbf{Beams}&\textbf{Columns} (cm) \\ 
\midrule
   $1- 20$& $W 24 \times192$&$50^2 \times 2.5$ \\
   $21-30$& $W 24 \times 103$&$40^2 \times 2.0$ \\
   $31-37$& $W 24 \times 103$&$35^2 \times 1.8$\\
 
\bottomrule
\end{tabular*}
\end{table}

\subsubsection{Wind load}
In this case, a non-directional mean hourly wind speed at the building top $v_H = 59.87$ m/s is considered. This makes the building go through noticeable nonlinearity, leading to a challenging case for metamodeling. Stochastic wind excitation is subsequently generated through the wind tunnel data-driven proper orthogonal decomposition model \cite{chen2005proper}. This was done by using the aerodynamic data from Tokyo Polytechnic University database, specifically a 1/300 scale rigid model tested at a mean reduced-scale top wind speed of 11m/s. The external wind pressures were recorded via 512 synchronous taps at a sampling frequency of 1000Hz, and were further processed into two translational floor loads and one torsional moment. The first five spatial modes were preserved at each discrete frequency point, and with a cutoff frequency of $1 Hz$ to reconstruct stochastic wind loads for the building. This wind load model was then leveraged to generate stationary wind excitation with a duration of 600 s and with a time step $\Delta t = 0.5$ s. Figure~\ref{fig:case3Structure} (b) and (c) are respectively the time histories and power spectral densities of the generated wind load, against those of the wind tunnel data at the top floor. Linear interpolation was used to adapt the dynamic wind load to finer time steps in direct integration for structural analysis. It should be noted that the wind tunnel data is only for the second quadrant, i.e., wind direction $\alpha$ = $0\degree$ to $90\degree$. A symmetric mapping from the second quadrant to the rest directions is thus performed \cite{li2024deep} to overcome this limitation. Since the wind directions at third and fourth quadrants completely mirrored the wind tunnel data from the second and the first quadrants, it is therefore only necessary to validate the applicability within the second and the first quadrants, i.e., the wind directions $\alpha$ = $0\degree$ to $90\degree$, and $270\degree$ to $350\degree$.

\subsubsection{Modeling and data generation}
A high-fidelity finite element model for the structural system was built in the OpenSees software version 3.7.1 \cite{OpenSees2000}. The structural components were modeled using displacement-based elements with fiber-discretized sections to capture distributed and progressive inelasticity. The Gauss-Legendre integration scheme with 5 integration points per element along the length was considered. Within each fiber, the elastic-perfectly-plastic steel material model was considered. The inherent damping within the system was modeled as random Rayleigh damping. This is done by introducing a random damping factor $\eta$ following a lognormal distribution, with the mean of 1.0 and the coefficient of variation of 0.3. The coefficients of the Rayleigh damping were first calibrated such that the deterministic modal damping ratios at the first two frequencies, 0.28 Hz and 0.81 Hz, are 2.5\%, then were multiplied by the random damping factor $\eta$.

The structural responses were solved with the average constant acceleration method. At each time step, the nonlinear equilibrium was solved by the adaptive nonlinear solver as in \cite{li2022metamodeling}. Following the successful execution of the OpenSees simulation, the Tcl scripts extract displacement response time histories and fiber hysteresis curves for each realization of wind excitation and uncertain parameter provided to the structure. The OpenSees Tcl scripts were executed in the MATLAB environment to loop over all realizations. 

POD-based dimensionality reduction and wavelet-based downsampling were carried out for the wind excitation and displacement responses. To perform the POD-based dimensionality reduction, 800 samples were set aside from the 1000 data points as the calibration dataset, which will be used for metamodel calibration in the next Section. The POD mode vectors were then extracted by performing singular value decomposition over displacement snapshots. The snapshots were collected by considering 1200 equally spaced time points from each displacement time history, leading to a collection of 800 $\times$ 1200 snapshots. By considering a threshold of the ratio between the cumulative summation and the total summation of squared singular values, i.e., the fraction of captured signal energy, of 99.99\% , the first three significant POD mode vectors were reserved as the POD basis. As such, the high-dimensional wind excitation and displacement responses were reduced to a space with a low dimensionality of three. Next, both the reduced inputs and outputs are then normalized by the maximum and minimum values obtained from the entire calibration set \cite{wang2020knowledge,li2022metamodeling}. Finally, the wavelet transformation was implemented by considering the 6th-order Daubechies function to further downsample the reduced inputs and outputs. This converts the reduced inputs and outputs to shortened wavelet coefficient series. The wavelet coefficient series of reduced inputs is then paired with the replicated random variable of the system, i.e., the random damping factor, as the input. The metamodel was calibrated to reproduce, based on the augmented input, the wavelet coefficient series of the reduced outputs. Once obtained, inverse wavelet transformation and POD modal superpositioning can be quickly performed to obtain the displacement and other responses. 

\subsubsection{Training and testing configurations}
\begin{figure}
    \centering
    \includegraphics[width=0.5\linewidth]{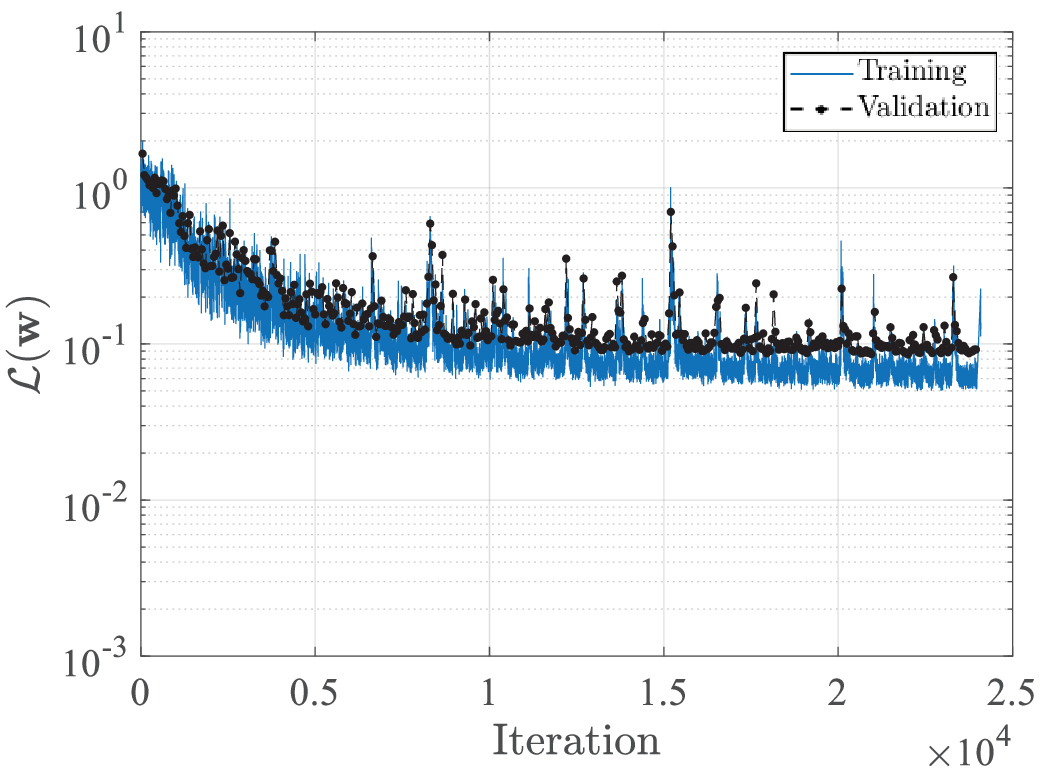}
    \caption{Training and validation loss of metamodel in Case 3}
    \label{fig:Case3Training}
\end{figure}

A metamodel of the same architecture as Cases 1 and 2, a LSTM network with an input layer, an LSTM layer with 200 hidden units, and a fully connected layer, was considered in this case. The metamodel takes the augmented inputs, consisting of the wavelet coefficient series of reduced inputs (representing wind load) and the replicated random damping factor, as input, and reproduces the wavelet coefficients of reduced outputs (which can further reconstruct displacement time history). The Monte Carlo dropout with a dropout rate of 20\% was implemented for the fully connected layer to account for the epistemic uncertainty through the variational inference. The calibration set was partitioned into training and validation sets using a 750:50 ratio. The rest of 200 samples that were not included in the calibration set are used, as a testing set, to test the model performance over unseen samples. The LSTM is trained using a mini-batch size of 50 and a learning rate of 0.005. The training phase is completed in 2400 epochs, reaching a total of $2.5 \times 10^4$ iterations. The curves of training and validation loss are shown in Figure~\ref{fig:Case3Training}. The training progress is completed in around 70 minutes. It is seen from the Figure~\ref{fig:Case3Training} that, although the loss curves exhibited noticeable fluctuations, they demonstrate an overall decreasing trend without apparent overfitting.

\subsubsection{Performance on the testing set}
\begin{figure}
    \centering
    \includegraphics[width=0.7\linewidth]{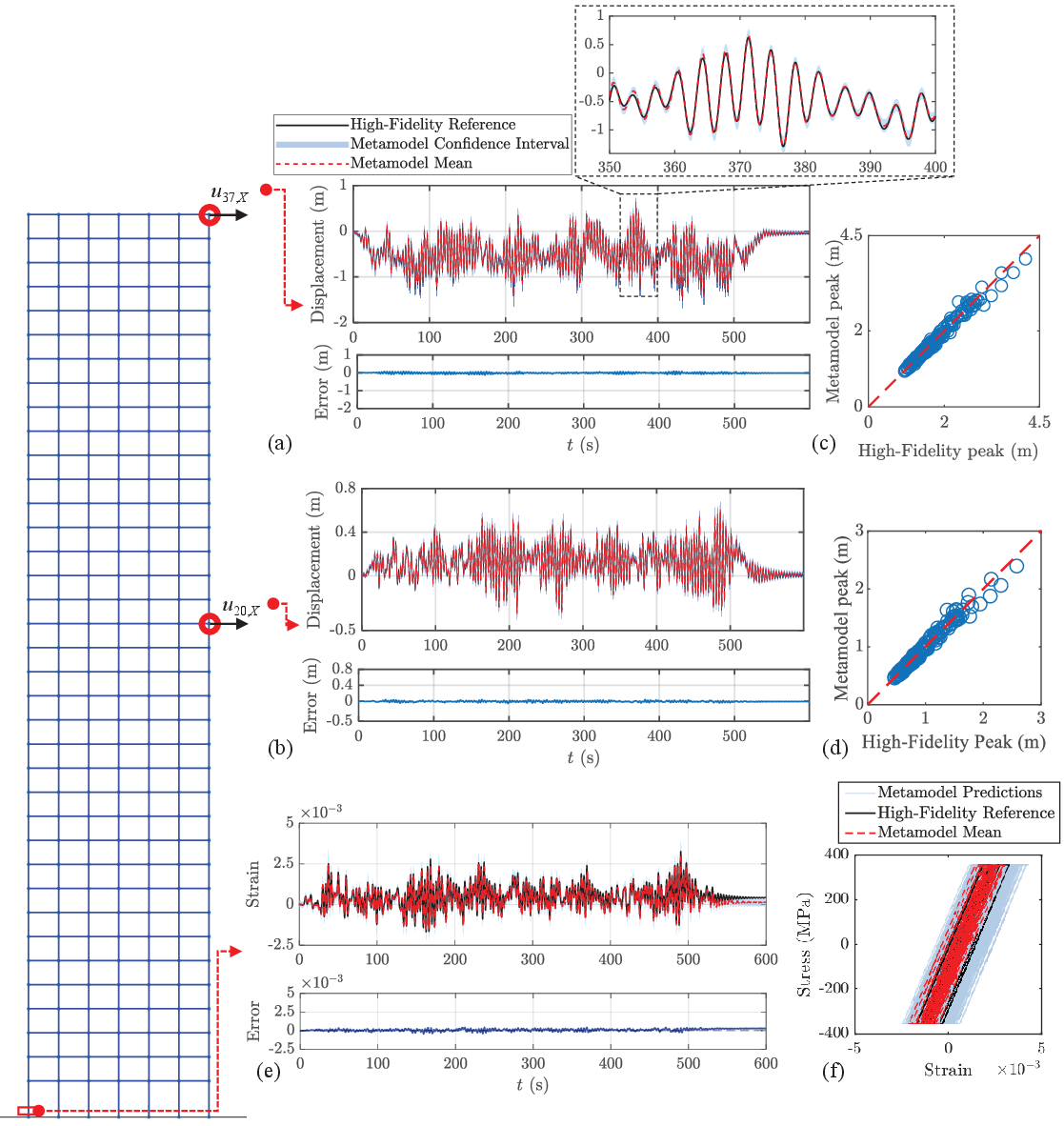}
    \caption{Comparisons of displacement time histories by the metamodel with 95\% of confidence intervals against those by the high-fidelity direct integration at 20th (a) and 37th floor (b); Peak displacements by the metamodel compared to those by the high-fidelity direct integration at 20th (c) and 37th floor (d); (e) Comparisons of strain time histories by the metamodel with 95\% of confidence intervals against those by the high-fidelity direct integration; (f) Comparison of the hysteresis curve by the metamodel, its 100 realizations under epistemic uncertainty, against those by the high-fidelity direct integration of an exterior fiber in the exterior column at the 1st floor.}
    \label{fig:Case3Results}
\end{figure}

Figure \ref{fig:Case3Results} shows the results of the metamodel validated on the 200 testing samples. The calibrated model in Case 3 is highly accurate with minimal discrepancy between the high-fidelity model and metamodel response, and exhibits different confidence between global and local responses. Figure~\ref{fig:Case3Results} (a) and (b) respectively show the comparisons of displacement time histories by the metamodel with 95\% confidence intervals against those by the high-fidelity direct integration at the 20th and 37th floor. The displacement response time histories by the metamodel exhibit an excellent match with those by the high-fidelity direct integration. This is also attested by the nearly zero error time histories in the scale of the responses. The high accuracy over the entire testing set is evidenced by the peak displacements by the metamodel compared to those by the high-fidelity direct integration at the 20th and 37th floor, respectively shown in Figure~\ref{fig:Case3Results} (c) and (d), where the data points overlap with the line of identity. Moreover, through the Monte Carlo dropout scheme, the epistemic uncertainty in response time histories is shown as the intervals of 95\% confidence in Figure~\ref{fig:Case3Results} (a) and (b), indicating the uncertainty arising from finite training data and effort. Once global displacements are obtained, local responses, including strain time histories and hyteresis curves, can be rapidly obtained as well. Figure~\ref{fig:Case3Results} (e) compares the strain time histories of an exterior fiber in the exterior column at the 1st floor (95\% confidence interval is included), calculated by the displacements from the metamodel and by the direct integration. Figure~\ref{fig:Case3Results} (f) shows the comparison of the hysteresis curve of the exterior fiber by the metamodel, its 100 realizations under epistemic uncertainty, against those by the high-fidelity direct integration. Notably, the epistemic uncertainty for these local responses (strain and hysteresis) is significantly higher than that observed for global displacements. This is because local responses are reproduced by displacement responses, which, in general, are usually sensitive to perturbations in displacement response time histories. This is consistent with the fact that, in metamodeling, local responses, including strain, stress, and hysteresis curves, are more challenging to accurately reproduce. This cannot be revealed if epistemic uncertainty is not explicitly considered.



\section{Conclusion}
In this paper, a metamodeling technique capable of propagating the full range of aleatoric and epistemic uncertainty was developed for high-dimensional nonlinear structural dynamic systems under stochastic excitations. The long short-term memory (LSTM) network is leveraged along with proper orthogonal decomposition (POD)-based dimensionality reduction and wavelet-based downsampling, as a sequence-to-sequence mapping to simulate a full range of displacement responses given stochastic excitation. To incorporate uncertainties within structural systems, the random parameters describing system uncertainty are considered as augmented inputs to the LSTM network. This allows for accurately capturing the effects of system uncertainty while retaining a one-to-one mapping from the random parameters of the system to responses. In addition, the epistemic uncertainty was effectively estimated by the Monte Carlo dropout scheme. Through a binary mask applied to the network parameters, the calibrated LSTM metamodel can generate an ensemble of realizations, and a confidence interval that represents the remaining epistemic uncertainty due to a finite amount of data and training effort, in addition to a point (mean) prediction of the response. 

The applicability of the developed scheme was validated comprehensively through three case studies involving an SDOF Bouc-Wen system subjected to stochastic seismic excitation, a shear building model under stochastic seismic excitation, and a full-scale steel moment-resisting framework subjected to stochastic wind excitation. All the cases are with system uncertainty from materials to dynamic properties. It was observed that the proposed LSTM network with augmented inputs accurately reproduced the structural responses under a full range of aleatoric uncertainty. On the other hand, the epistemic uncertainty can be estimated by the Monte Carlo dropout scheme, rendered as confidence intervals or response realizations. The epistemic uncertainty provides insights into the remaining uncertainty in the metamodel after calibration, providing a quantitative description of the additional risk from a finite amount of data and training effort. It is also interesting to observe that when reproducing local responses, including strain, stress, and hysteresis curves of a section fiber in a full-scale structure, higher epistemic uncertainty was observed. This observation aligns with the inherent challenges of accurately reproducing localized nonlinear behavior via metamodeling. Overall, the developed metamodeling techniques based on variational LSTM with augmented inputs enable the propagation of a full range of aleatoric and epistemic uncertainty, showing great potential in risk-informed design and engineering.

\section*{Data availability}
All codes will made available on GitHub\footnote{\url{https://github.com/jacklbw/Variational-LSTM-with-Augmented-Inputs}}, upon publication of the paper.

\section*{Acknowledgements}
The authors’ research efforts were partly supported by the faculty startup funding at Texas Tech University and Rensselaer Polytechnic Institute, and the Koh Family Scholarship at Texas Tech University. Any opinions, findings, conclusions, or recommendations expressed in this material are those of the author(s) and do not necessarily reflect the views of the funding organizations.


\bibliographystyle{cas-model2-names}

\bibliography{cas-refs}

\begin{thebibliography}{75}
\expandafter\ifx\csname natexlab\endcsname\relax\def\natexlab#1{#1}\fi
\providecommand{\url}[1]{\texttt{#1}}
\providecommand{\href}[2]{#2}
\providecommand{\path}[1]{#1}
\providecommand{\DOIprefix}{doi:}
\providecommand{\ArXivprefix}{arXiv:}
\providecommand{\URLprefix}{URL: }
\providecommand{\Pubmedprefix}{pmid:}
\providecommand{\doi}[1]{\href{http://dx.doi.org/#1}{\path{#1}}}
\providecommand{\Pubmed}[1]{\href{pmid:#1}{\path{#1}}}
\providecommand{\bibinfo}[2]{#2}
\ifx\xfnm\relax \def\xfnm[#1]{\unskip,\space#1}\fi
\bibitem[{Augusti and Ciampoli(2008)}]{augusti2008performance}
\bibinfo{author}{Augusti, G.}, \bibinfo{author}{Ciampoli, M.},
  \bibinfo{year}{2008}.
\newblock \bibinfo{title}{Performance-based design in risk assessment and
  reduction}.
\newblock \bibinfo{journal}{Probabilistic Engineering Mechanics}
  \bibinfo{volume}{23}, \bibinfo{pages}{496--508}.
\bibitem[{Baker and Cornell(2008)}]{baker2008uncertainty}
\bibinfo{author}{Baker, J.W.}, \bibinfo{author}{Cornell, C.A.},
  \bibinfo{year}{2008}.
\newblock \bibinfo{title}{Uncertainty propagation in probabilistic seismic loss
  estimation}.
\newblock \bibinfo{journal}{Structural Safety} \bibinfo{volume}{30},
  \bibinfo{pages}{236--252}.
\bibitem[{Barbato et~al.(2013)Barbato, Petrini, Unnikrishnan and
  Ciampoli}]{barbato2013performance}
\bibinfo{author}{Barbato, M.}, \bibinfo{author}{Petrini, F.},
  \bibinfo{author}{Unnikrishnan, V.U.}, \bibinfo{author}{Ciampoli, M.},
  \bibinfo{year}{2013}.
\newblock \bibinfo{title}{Performance-based hurricane engineering (pbhe)
  framework}.
\newblock \bibinfo{journal}{Structural safety} \bibinfo{volume}{45},
  \bibinfo{pages}{24--35}.
\bibitem[{Bernardini et~al.(2015)Bernardini, Spence, Kwon and
  Kareem}]{bernardini2015performance}
\bibinfo{author}{Bernardini, E.}, \bibinfo{author}{Spence, S.M.},
  \bibinfo{author}{Kwon, D.K.}, \bibinfo{author}{Kareem, A.},
  \bibinfo{year}{2015}.
\newblock \bibinfo{title}{Performance-based design of high-rise buildings for
  occupant comfort}.
\newblock \bibinfo{journal}{Journal of Structural Engineering}
  \bibinfo{volume}{141}, \bibinfo{pages}{04014244}.
\bibitem[{Bezabeh et~al.(2021)Bezabeh, Bitsuamlak and
  Tesfamariam}]{bezabeh2021nonlinear}
\bibinfo{author}{Bezabeh, M.A.}, \bibinfo{author}{Bitsuamlak, G.T.},
  \bibinfo{author}{Tesfamariam, S.}, \bibinfo{year}{2021}.
\newblock \bibinfo{title}{Nonlinear dynamic response of
  single-degree-of-freedom systems subjected to along-wind loads. i: Parametric
  study}.
\newblock \bibinfo{journal}{Journal of Structural Engineering}
  \bibinfo{volume}{147}, \bibinfo{pages}{04021177}.
\bibitem[{Bhattacharyya et~al.(2020)Bhattacharyya, Jacquelin and
  Brizard}]{Bhattacharyya20}
\bibinfo{author}{Bhattacharyya, B.}, \bibinfo{author}{Jacquelin, E.},
  \bibinfo{author}{Brizard, D.}, \bibinfo{year}{2020}.
\newblock \bibinfo{title}{A {K}riging–{NARX} model for uncertainty
  quantification of nonlinear stochastic dynamical systems in time domain}.
\newblock \bibinfo{journal}{J. Eng. Mech.} \bibinfo{volume}{146},
  \bibinfo{pages}{04020070}.
\bibitem[{Chatterjee and Chowdhury(2017)}]{chatterjee2017efficient}
\bibinfo{author}{Chatterjee, T.}, \bibinfo{author}{Chowdhury, R.},
  \bibinfo{year}{2017}.
\newblock \bibinfo{title}{An efficient sparse bayesian learning framework for
  stochastic response analysis}.
\newblock \bibinfo{journal}{Structural Safety} \bibinfo{volume}{68},
  \bibinfo{pages}{1--14}.
\bibitem[{Chen and Kareem(2005)}]{chen2005proper}
\bibinfo{author}{Chen, X.}, \bibinfo{author}{Kareem, A.}, \bibinfo{year}{2005}.
\newblock \bibinfo{title}{Proper orthogonal decomposition-based modeling,
  analysis, and simulation of dynamic wind load effects on structures}.
\newblock \bibinfo{journal}{Journal of Engineering Mechanics}
  \bibinfo{volume}{131}, \bibinfo{pages}{325--339}.
\bibitem[{Chuang and Spence(2017)}]{chuang2017performance}
\bibinfo{author}{Chuang, W.C.}, \bibinfo{author}{Spence, S.M.},
  \bibinfo{year}{2017}.
\newblock \bibinfo{title}{A performance-based design framework for the
  integrated collapse and non-collapse assessment of wind excited buildings}.
\newblock \bibinfo{journal}{Engineering Structures} \bibinfo{volume}{150},
  \bibinfo{pages}{746--758}.
\bibitem[{Chuang and Spence(2022)}]{chuang2022framework}
\bibinfo{author}{Chuang, W.C.}, \bibinfo{author}{Spence, S.M.},
  \bibinfo{year}{2022}.
\newblock \bibinfo{title}{A framework for the efficient reliability assessment
  of inelastic wind excited structures at dynamic shakedown}.
\newblock \bibinfo{journal}{Journal of Wind Engineering and Industrial
  Aerodynamics} \bibinfo{volume}{220}, \bibinfo{pages}{104834}.
\bibitem[{Ciampoli et~al.(2011)Ciampoli, Petrini and
  Augusti}]{ciampoli2011performance}
\bibinfo{author}{Ciampoli, M.}, \bibinfo{author}{Petrini, F.},
  \bibinfo{author}{Augusti, G.}, \bibinfo{year}{2011}.
\newblock \bibinfo{title}{Performance-based wind engineering: towards a general
  procedure}.
\newblock \bibinfo{journal}{Structural safety} \bibinfo{volume}{33},
  \bibinfo{pages}{367--378}.
\bibitem[{Cui and Caracoglia(2018)}]{cui2018unified}
\bibinfo{author}{Cui, W.}, \bibinfo{author}{Caracoglia, L.},
  \bibinfo{year}{2018}.
\newblock \bibinfo{title}{A unified framework for performance-based wind
  engineering of tall buildings in hurricane-prone regions based on lifetime
  intervention-cost estimation}.
\newblock \bibinfo{journal}{Structural safety} \bibinfo{volume}{73},
  \bibinfo{pages}{75--86}.
\bibitem[{Cui and Caracoglia(2020)}]{cui2020performance}
\bibinfo{author}{Cui, W.}, \bibinfo{author}{Caracoglia, L.},
  \bibinfo{year}{2020}.
\newblock \bibinfo{title}{Performance-based wind engineering of tall buildings
  examining life-cycle downtime and multisource wind damage}.
\newblock \bibinfo{journal}{Journal of Structural Engineering}
  \bibinfo{volume}{146}, \bibinfo{pages}{04019179}.
\bibitem[{De~Grandis et~al.(2009)De~Grandis, Domaneschi and
  Perotti}]{de2009numerical}
\bibinfo{author}{De~Grandis, S.}, \bibinfo{author}{Domaneschi, M.},
  \bibinfo{author}{Perotti, F.}, \bibinfo{year}{2009}.
\newblock \bibinfo{title}{A numerical procedure for computing the fragility of
  npp components under random seismic excitation}.
\newblock \bibinfo{journal}{Nuclear Engineering and Design}
  \bibinfo{volume}{239}, \bibinfo{pages}{2491--2499}.
\bibitem[{Der~Kiureghian and Ditlevsen(2009)}]{der2009aleatory}
\bibinfo{author}{Der~Kiureghian, A.}, \bibinfo{author}{Ditlevsen, O.},
  \bibinfo{year}{2009}.
\newblock \bibinfo{title}{Aleatory or epistemic? does it matter?}
\newblock \bibinfo{journal}{Structural safety} \bibinfo{volume}{31},
  \bibinfo{pages}{105--112}.
\bibitem[{Ding et~al.(2023)Ding, Feng, Brunesi, Parisi and
  Wu}]{ding2023efficient}
\bibinfo{author}{Ding, J.Y.}, \bibinfo{author}{Feng, D.C.},
  \bibinfo{author}{Brunesi, E.}, \bibinfo{author}{Parisi, F.},
  \bibinfo{author}{Wu, G.}, \bibinfo{year}{2023}.
\newblock \bibinfo{title}{Efficient seismic fragility analysis method utilizing
  ground motion clustering and probabilistic machine learning}.
\newblock \bibinfo{journal}{Engineering Structures} \bibinfo{volume}{294},
  \bibinfo{pages}{116739}.
\bibitem[{Ellingwood(2008)}]{ellingwood2008structural}
\bibinfo{author}{Ellingwood, B.R.}, \bibinfo{year}{2008}.
\newblock \bibinfo{title}{Structural reliability and performance-based
  engineering}.
\newblock \bibinfo{journal}{Proceedings of the Institution of Civil
  Engineers-Structures and Buildings} \bibinfo{volume}{161},
  \bibinfo{pages}{199--207}.
\bibitem[{Gal and Ghahramani(2015)}]{gal2015dropout}
\bibinfo{author}{Gal, Y.}, \bibinfo{author}{Ghahramani, Z.},
  \bibinfo{year}{2015}.
\newblock \bibinfo{title}{Dropout as a bayesian approximation}.
\newblock \bibinfo{journal}{arXiv preprint arXiv:1506.02157} .
\bibitem[{Gal and Ghahramani(2016)}]{gal2016theoretically}
\bibinfo{author}{Gal, Y.}, \bibinfo{author}{Ghahramani, Z.},
  \bibinfo{year}{2016}.
\newblock \bibinfo{title}{A theoretically grounded application of dropout in
  recurrent neural networks}.
\newblock \bibinfo{journal}{Advances in neural information processing systems}
  \bibinfo{volume}{29}.
\bibitem[{Gal et~al.(2016)}]{gal2016uncertainty}
\bibinfo{author}{Gal, Y.}, et~al., \bibinfo{year}{2016}.
\newblock \bibinfo{title}{Uncertainty in deep learning} .
\bibitem[{Gardoni and LaFave(2016)}]{gardoni2016multi}
\bibinfo{author}{Gardoni, P.}, \bibinfo{author}{LaFave, J.M.},
  \bibinfo{year}{2016}.
\newblock \bibinfo{title}{Multi-hazard approaches to civil infrastructure
  engineering: Mitigating risks and promoting resilence}, in:
  \bibinfo{booktitle}{Multi-hazard approaches to civil infrastructure
  engineering}. \bibinfo{publisher}{Springer}, pp. \bibinfo{pages}{3--12}.
\bibitem[{Ghosh and Chakraborty(2020)}]{ghosh2020seismic}
\bibinfo{author}{Ghosh, S.}, \bibinfo{author}{Chakraborty, S.},
  \bibinfo{year}{2020}.
\newblock \bibinfo{title}{Seismic fragility analysis of structures based on
  bayesian linear regression demand models}.
\newblock \bibinfo{journal}{Probabilistic Engineering Mechanics}
  \bibinfo{volume}{61}, \bibinfo{pages}{103081}.
\bibitem[{Ghosh et~al.(2019)Ghosh, Roy and Chakraborty}]{ghosh2019kriging}
\bibinfo{author}{Ghosh, S.}, \bibinfo{author}{Roy, A.},
  \bibinfo{author}{Chakraborty, S.}, \bibinfo{year}{2019}.
\newblock \bibinfo{title}{Kriging metamodeling-based monte carlo simulation for
  improved seismic fragility analysis of structures}.
\newblock \bibinfo{journal}{Journal of Earthquake Engineering} ,
  \bibinfo{pages}{1--21}.
\bibitem[{Gidaris et~al.(2015)Gidaris, Taflanidis and
  Mavroeidis}]{gidaris2015kriging}
\bibinfo{author}{Gidaris, I.}, \bibinfo{author}{Taflanidis, A.A.},
  \bibinfo{author}{Mavroeidis, G.P.}, \bibinfo{year}{2015}.
\newblock \bibinfo{title}{Kriging metamodeling in seismic risk assessment based
  on stochastic ground motion models}.
\newblock \bibinfo{journal}{Earthquake Engineering \& Structural Dynamics}
  \bibinfo{volume}{44}, \bibinfo{pages}{2377--2399}.
\bibitem[{Goswami et~al.(2025)Goswami, Giovanis, Li, Spence and
  Shields}]{GOSWAMI2025121284}
\bibinfo{author}{Goswami, S.}, \bibinfo{author}{Giovanis, D.G.},
  \bibinfo{author}{Li, B.}, \bibinfo{author}{Spence, S.M.},
  \bibinfo{author}{Shields, M.D.}, \bibinfo{year}{2025}.
\newblock \bibinfo{title}{Neural operators for stochastic modeling of nonlinear
  structural system response to natural hazards}.
\newblock \bibinfo{journal}{Engineering Structures} \bibinfo{volume}{345},
  \bibinfo{pages}{121284}.
\newblock \URLprefix
  \url{https://www.sciencedirect.com/science/article/pii/S014102962501675X},
  \DOIprefix\doi{https://doi.org/10.1016/j.engstruct.2025.121284}.
\bibitem[{G{\"u}nay and Mosalam(2013)}]{gunay2013peer}
\bibinfo{author}{G{\"u}nay, S.}, \bibinfo{author}{Mosalam, K.M.},
  \bibinfo{year}{2013}.
\newblock \bibinfo{title}{Peer performance-based earthquake engineering
  methodology, revisited}.
\newblock \bibinfo{journal}{Journal of Earthquake Engineering}
  \bibinfo{volume}{17}, \bibinfo{pages}{829--858}.
\bibitem[{Hinton et~al.(2012)Hinton, Srivastava, Krizhevsky, Sutskever and
  Salakhutdinov}]{hinton2012improving}
\bibinfo{author}{Hinton, G.E.}, \bibinfo{author}{Srivastava, N.},
  \bibinfo{author}{Krizhevsky, A.}, \bibinfo{author}{Sutskever, I.},
  \bibinfo{author}{Salakhutdinov, R.R.}, \bibinfo{year}{2012}.
\newblock \bibinfo{title}{Improving neural networks by preventing co-adaptation
  of feature detectors}.
\newblock \bibinfo{journal}{arXiv preprint arXiv:1207.0580} .
\bibitem[{Hong(2004)}]{hong2004accumulation}
\bibinfo{author}{Hong, H.}, \bibinfo{year}{2004}.
\newblock \bibinfo{title}{Accumulation of wind induced damage on bilinear sdof
  systems}.
\newblock \bibinfo{journal}{Wind \& structures} \bibinfo{volume}{7},
  \bibinfo{pages}{145--158}.
\bibitem[{Huang and Chen(2022)}]{huang2022inelastic}
\bibinfo{author}{Huang, J.}, \bibinfo{author}{Chen, X.}, \bibinfo{year}{2022}.
\newblock \bibinfo{title}{Inelastic performance of high-rise buildings to
  simultaneous actions of alongwind and crosswind loads}.
\newblock \bibinfo{journal}{Journal of Structural Engineering}
  \bibinfo{volume}{148}, \bibinfo{pages}{04021258}.
\bibitem[{Ierimonti et~al.(2019)Ierimonti, Venanzi, Caracoglia and
  Materazzi}]{ierimonti2019cost}
\bibinfo{author}{Ierimonti, L.}, \bibinfo{author}{Venanzi, I.},
  \bibinfo{author}{Caracoglia, L.}, \bibinfo{author}{Materazzi, A.L.},
  \bibinfo{year}{2019}.
\newblock \bibinfo{title}{Cost-based design of nonstructural elements for tall
  buildings under extreme wind environments}.
\newblock \bibinfo{journal}{Journal of Aerospace Engineering}
  \bibinfo{volume}{32}, \bibinfo{pages}{04019020}.
\bibitem[{Judd and Charney(2015)}]{judd2015inelastic}
\bibinfo{author}{Judd, J.}, \bibinfo{author}{Charney, F.A.},
  \bibinfo{year}{2015}.
\newblock \bibinfo{title}{Inelastic behavior and collapse risk for buildings
  subjected to wind loads}, in: \bibinfo{booktitle}{Structures Congress 2015},
  pp. \bibinfo{pages}{2483--2496}.
\bibitem[{Kendall and Gal(2017)}]{kendall2017uncertainties}
\bibinfo{author}{Kendall, A.}, \bibinfo{author}{Gal, Y.}, \bibinfo{year}{2017}.
\newblock \bibinfo{title}{What uncertainties do we need in bayesian deep
  learning for computer vision?}
\newblock \bibinfo{journal}{Advances in neural information processing systems}
  \bibinfo{volume}{30}.
\bibitem[{Kim et~al.(2019)Kim, Kwon and Song}]{kim2019response}
\bibinfo{author}{Kim, T.}, \bibinfo{author}{Kwon, O.S.}, \bibinfo{author}{Song,
  J.}, \bibinfo{year}{2019}.
\newblock \bibinfo{title}{Response prediction of nonlinear hysteretic systems
  by deep neural networks}.
\newblock \bibinfo{journal}{Neural Networks} \bibinfo{volume}{111},
  \bibinfo{pages}{1--10}.
\bibitem[{Kim et~al.(2020)Kim, Song and Kwon}]{kim2020probabilistic}
\bibinfo{author}{Kim, T.}, \bibinfo{author}{Song, J.}, \bibinfo{author}{Kwon,
  O.S.}, \bibinfo{year}{2020}.
\newblock \bibinfo{title}{Probabilistic evaluation of seismic responses using
  deep learning method}.
\newblock \bibinfo{journal}{Structural Safety} \bibinfo{volume}{84},
  \bibinfo{pages}{101913}.
\bibitem[{Koutsourelakis(2010)}]{koutsourelakis2010assessing}
\bibinfo{author}{Koutsourelakis, P.S.}, \bibinfo{year}{2010}.
\newblock \bibinfo{title}{Assessing structural vulnerability against
  earthquakes using multi-dimensional fragility surfaces: A bayesian
  framework}.
\newblock \bibinfo{journal}{Probabilistic Engineering Mechanics}
  \bibinfo{volume}{25}, \bibinfo{pages}{49--60}.
\bibitem[{Kroetz et~al.(2017)Kroetz, Tessari and Beck}]{kroetz2017performance}
\bibinfo{author}{Kroetz, H.M.}, \bibinfo{author}{Tessari, R.K.},
  \bibinfo{author}{Beck, A.T.}, \bibinfo{year}{2017}.
\newblock \bibinfo{title}{Performance of global metamodeling techniques in
  solution of structural reliability problems}.
\newblock \bibinfo{journal}{Advances in Engineering Software}
  \bibinfo{volume}{114}, \bibinfo{pages}{394--404}.
\bibitem[{Kundu and Chakraborty(2020)}]{kundu2020deep}
\bibinfo{author}{Kundu, A.}, \bibinfo{author}{Chakraborty, S.},
  \bibinfo{year}{2020}.
\newblock \bibinfo{title}{Deep learning-based metamodeling technique for
  nonlinear seismic response quantification}, in: \bibinfo{booktitle}{IOP
  Conference Series: Materials Science and Engineering},
  \bibinfo{organization}{IOP Publishing}. p. \bibinfo{pages}{012042}.
\bibitem[{Kyprioti and Taflanidis(2021)}]{kyprioti2021kriging}
\bibinfo{author}{Kyprioti, A.P.}, \bibinfo{author}{Taflanidis, A.A.},
  \bibinfo{year}{2021}.
\newblock \bibinfo{title}{Kriging metamodeling for seismic response
  distribution estimation}.
\newblock \bibinfo{journal}{Earthquake Engineering \& Structural Dynamics}
  \bibinfo{volume}{50}, \bibinfo{pages}{3550--3576}.
\bibitem[{Le and Caracoglia(2015)}]{le2015reduced}
\bibinfo{author}{Le, T.H.}, \bibinfo{author}{Caracoglia, L.},
  \bibinfo{year}{2015}.
\newblock \bibinfo{title}{Reduced-order wavelet-galerkin solution for the
  coupled, nonlinear stochastic response of slender buildings in transient
  winds}.
\newblock \bibinfo{journal}{Journal of Sound and Vibration}
  \bibinfo{volume}{344}, \bibinfo{pages}{179--208}.
\bibitem[{Li et~al.(2021)Li, Chuang and Spence}]{li2021response}
\bibinfo{author}{Li, B.}, \bibinfo{author}{Chuang, W.C.},
  \bibinfo{author}{Spence, S.M.}, \bibinfo{year}{2021}.
\newblock \bibinfo{title}{Response estimation of multi-degree-of-freedom
  nonlinear stochastic structural systems through metamodeling}.
\newblock \bibinfo{journal}{Journal of Engineering Mechanics}
  \bibinfo{volume}{147}, \bibinfo{pages}{04021082}.
\bibitem[{Li and Spence(2022)}]{li2022metamodeling}
\bibinfo{author}{Li, B.}, \bibinfo{author}{Spence, S.M.}, \bibinfo{year}{2022}.
\newblock \bibinfo{title}{Metamodeling through deep learning of
  high-dimensional dynamic nonlinear systems driven by general stochastic
  excitation}.
\newblock \bibinfo{journal}{Journal of Structural Engineering}
  \bibinfo{volume}{148}, \bibinfo{pages}{04022186}.
\bibitem[{Li and Spence(2024)}]{li2024deep}
\bibinfo{author}{Li, B.}, \bibinfo{author}{Spence, S.M.}, \bibinfo{year}{2024}.
\newblock \bibinfo{title}{Deep learning enabled rapid nonlinear time history
  wind performance assessment}, in: \bibinfo{booktitle}{Structures},
  \bibinfo{organization}{Elsevier}. p. \bibinfo{pages}{106810}.
\bibitem[{Li et~al.(2024)Li, Yang, Deng, Gong, Tian, Su and Teng}]{li2024time}
\bibinfo{author}{Li, Z.}, \bibinfo{author}{Yang, Q.}, \bibinfo{author}{Deng,
  Q.}, \bibinfo{author}{Gong, Y.}, \bibinfo{author}{Tian, D.},
  \bibinfo{author}{Su, P.}, \bibinfo{author}{Teng, J.}, \bibinfo{year}{2024}.
\newblock \bibinfo{title}{Time history seismic response prediction of multiple
  homogeneous building structures using only one deep learning-based structure
  temporal fusion network}.
\newblock \bibinfo{journal}{Earthquake Engineering \& Structural Dynamics}
  \bibinfo{volume}{53}, \bibinfo{pages}{4076--4098}.
\bibitem[{Mai et~al.(2016)Mai, Spiridonakos, Chatzi and
  Sudret}]{mai2016surrogate}
\bibinfo{author}{Mai, C.V.}, \bibinfo{author}{Spiridonakos, M.D.},
  \bibinfo{author}{Chatzi, E.N.}, \bibinfo{author}{Sudret, B.},
  \bibinfo{year}{2016}.
\newblock \bibinfo{title}{Surrogate modeling for stochastic dynamical systems
  by combining nonlinear autoregressive with exogenous input models and
  polynomial chaos expansions}.
\newblock \bibinfo{journal}{International Journal for Uncertainty
  Quantification} \bibinfo{volume}{6}.
\bibitem[{McKenna et~al.(2000)McKenna, Fenves and Scott}]{OpenSees2000}
\bibinfo{author}{McKenna, F.}, \bibinfo{author}{Fenves, G.L.},
  \bibinfo{author}{Scott, M.H.}, \bibinfo{year}{2000}.
\newblock \bibinfo{title}{Open System for Earthquake Engineering Simulation}.
\newblock \bibinfo{organization}{University of California, Berkeley}.
\newblock \URLprefix \url{http://opensees.berkeley.edu}.
  \bibinfo{note}{\url{http://opensees.berkeley.edu}}.
\bibitem[{McKenna et~al.(2024)McKenna, Yi, Bangalore~Satish, Zsarnoczay,
  Gardner and Elhaddad}]{McKenna2024quoFEM}
\bibinfo{author}{McKenna, F.}, \bibinfo{author}{Yi, S.r.},
  \bibinfo{author}{Bangalore~Satish, A.}, \bibinfo{author}{Zsarnoczay, A.},
  \bibinfo{author}{Gardner, M.}, \bibinfo{author}{Elhaddad, W.},
  \bibinfo{year}{2024}.
\newblock \bibinfo{title}{{NHERI-SimCenter/quoFEM: Version 4.0.0}}.
\newblock \URLprefix \url{https://doi.org/10.5281/zenodo.10443180},
  \DOIprefix\doi{10.5281/zenodo.10443180}.
\bibitem[{Micheli et~al.(2019)Micheli, Alipour, Laflamme and
  Sarkar}]{micheli2019performance}
\bibinfo{author}{Micheli, L.}, \bibinfo{author}{Alipour, A.},
  \bibinfo{author}{Laflamme, S.}, \bibinfo{author}{Sarkar, P.},
  \bibinfo{year}{2019}.
\newblock \bibinfo{title}{Performance-based design with life-cycle cost
  assessment for damping systems integrated in wind excited tall buildings}.
\newblock \bibinfo{journal}{Engineering Structures} \bibinfo{volume}{195},
  \bibinfo{pages}{438--451}.
\bibitem[{Moehle and Deierlein(2004)}]{moehle2004framework}
\bibinfo{author}{Moehle, J.}, \bibinfo{author}{Deierlein, G.G.},
  \bibinfo{year}{2004}.
\newblock \bibinfo{title}{A framework methodology for performance-based
  earthquake engineering}, in: \bibinfo{booktitle}{13th world conference on
  earthquake engineering}, \bibinfo{organization}{WCEE Vancouver}.
  p.~\bibinfo{pages}{12}.
\bibitem[{Ning and Xie(2024)}]{ning2024dual}
\bibinfo{author}{Ning, C.}, \bibinfo{author}{Xie, Y.}, \bibinfo{year}{2024}.
\newblock \bibinfo{title}{Dual-branch deep learning model for seismic response
  history prediction of bridge portfolios}, in: \bibinfo{booktitle}{18th World
  Conference on Earthquake Engineering}, \bibinfo{organization}{World
  Conference on Earthquake Engineering (WCEE)}.
\newblock \bibinfo{note}{Conference paper presented in Milan, Italy}.
\bibitem[{Ouyang and Spence(2020)}]{ouyang2020performance}
\bibinfo{author}{Ouyang, Z.}, \bibinfo{author}{Spence, S.M.},
  \bibinfo{year}{2020}.
\newblock \bibinfo{title}{A performance-based wind engineering framework for
  envelope systems of engineered buildings subject to directional wind and rain
  hazards}.
\newblock \bibinfo{journal}{Journal of Structural Engineering}
  \bibinfo{volume}{146}, \bibinfo{pages}{04020049}.
\bibitem[{Peng et~al.(2026)Peng, Taflanidis and Zhang}]{peng2026accelerating}
\bibinfo{author}{Peng, H.}, \bibinfo{author}{Taflanidis, A.A.},
  \bibinfo{author}{Zhang, J.}, \bibinfo{year}{2026}.
\newblock \bibinfo{title}{Accelerating seismic response distribution estimation
  with scalable mixture density network stochastic surrogate models}.
\newblock \bibinfo{journal}{Earthquake Engineering \& Structural Dynamics}
  \bibinfo{volume}{55}, \bibinfo{pages}{721--742}.
\bibitem[{Perotti et~al.(2013)Perotti, Domaneschi and
  De~Grandis}]{perotti2013numerical}
\bibinfo{author}{Perotti, F.}, \bibinfo{author}{Domaneschi, M.},
  \bibinfo{author}{De~Grandis, S.}, \bibinfo{year}{2013}.
\newblock \bibinfo{title}{The numerical computation of seismic fragility of
  base-isolated nuclear power plants buildings}.
\newblock \bibinfo{journal}{Nuclear Engineering and Design}
  \bibinfo{volume}{262}, \bibinfo{pages}{189--200}.
\bibitem[{Pujari et~al.(2016)Pujari, Ghosh and Lala}]{pujari2016bayesian}
\bibinfo{author}{Pujari, N.N.}, \bibinfo{author}{Ghosh, S.},
  \bibinfo{author}{Lala, S.}, \bibinfo{year}{2016}.
\newblock \bibinfo{title}{Bayesian approach for the seismic fragility
  estimation of a containment shell based on the formation of through-wall
  cracks}.
\newblock \bibinfo{journal}{ASCE-ASME Journal of Risk and Uncertainty in
  Engineering Systems, Part A: Civil Engineering} \bibinfo{volume}{2},
  \bibinfo{pages}{B4015004}.
\bibitem[{Rezaeian and {Der Kiureghian}(2010)}]{rezaeian2010}
\bibinfo{author}{Rezaeian, S.}, \bibinfo{author}{{Der Kiureghian}, A.},
  \bibinfo{year}{2010}.
\newblock \bibinfo{title}{Simulation of synthetic ground motions for specified
  earthquake and site characteristics}.
\newblock \bibinfo{journal}{Earthq. Eng. Struct. Dyn.} \bibinfo{volume}{39},
  \bibinfo{pages}{1155--1180}.
\bibitem[{Saha et~al.(2016)Saha, Matsagar and
  Chakraborty}]{saha2016uncertainty}
\bibinfo{author}{Saha, S.K.}, \bibinfo{author}{Matsagar, V.},
  \bibinfo{author}{Chakraborty, S.}, \bibinfo{year}{2016}.
\newblock \bibinfo{title}{Uncertainty quantification and seismic fragility of
  base-isolated liquid storage tanks using response surface models}.
\newblock \bibinfo{journal}{Probabilistic Engineering Mechanics}
  \bibinfo{volume}{43}, \bibinfo{pages}{20--35}.
\bibitem[{Segura et~al.(2020)Segura, Padgett and Paultre}]{segura2020metamodel}
\bibinfo{author}{Segura, R.}, \bibinfo{author}{Padgett, J.E.},
  \bibinfo{author}{Paultre, P.}, \bibinfo{year}{2020}.
\newblock \bibinfo{title}{Metamodel-based seismic fragility analysis of
  concrete gravity dams}.
\newblock \bibinfo{journal}{Journal of Structural Engineering}
  \bibinfo{volume}{146}, \bibinfo{pages}{04020121}.
\bibitem[{Seo and Caracoglia(2013)}]{seo2013estimating}
\bibinfo{author}{Seo, D.W.}, \bibinfo{author}{Caracoglia, L.},
  \bibinfo{year}{2013}.
\newblock \bibinfo{title}{Estimating life-cycle monetary losses due to wind
  hazards: Fragility analysis of long-span bridges}.
\newblock \bibinfo{journal}{Engineering Structures} \bibinfo{volume}{56},
  \bibinfo{pages}{1593--1606}.
\bibitem[{Seo et~al.(2012)Seo, Due{\~n}as-Osorio, Craig and
  Goodno}]{seo2012metamodel}
\bibinfo{author}{Seo, J.}, \bibinfo{author}{Due{\~n}as-Osorio, L.},
  \bibinfo{author}{Craig, J.I.}, \bibinfo{author}{Goodno, B.J.},
  \bibinfo{year}{2012}.
\newblock \bibinfo{title}{Metamodel-based regional vulnerability estimate of
  irregular steel moment-frame structures subjected to earthquake events}.
\newblock \bibinfo{journal}{Engineering Structures} \bibinfo{volume}{45},
  \bibinfo{pages}{585--597}.
\bibitem[{Shi et~al.(2019)Shi, Lu, Xu and Chen}]{shi2019adaptive}
\bibinfo{author}{Shi, Y.}, \bibinfo{author}{Lu, Z.}, \bibinfo{author}{Xu, L.},
  \bibinfo{author}{Chen, S.}, \bibinfo{year}{2019}.
\newblock \bibinfo{title}{An adaptive multiple-kriging-surrogate method for
  time-dependent reliability analysis}.
\newblock \bibinfo{journal}{Applied Mathematical Modelling}
  \bibinfo{volume}{70}, \bibinfo{pages}{545--571}.
\bibitem[{Simpson et~al.(2021)Simpson, Dervilis and
  Chatzi}]{simpson2021machine}
\bibinfo{author}{Simpson, T.}, \bibinfo{author}{Dervilis, N.},
  \bibinfo{author}{Chatzi, E.}, \bibinfo{year}{2021}.
\newblock \bibinfo{title}{Machine learning approach to model order reduction of
  nonlinear systems via autoencoder and lstm networks}.
\newblock \bibinfo{journal}{Journal of Engineering Mechanics}
  \bibinfo{volume}{147}, \bibinfo{pages}{04021061}.
\bibitem[{Spence and Kareem(2014)}]{spence2014performance}
\bibinfo{author}{Spence, S.M.}, \bibinfo{author}{Kareem, A.},
  \bibinfo{year}{2014}.
\newblock \bibinfo{title}{Performance-based design and optimization of
  uncertain wind-excited dynamic building systems}.
\newblock \bibinfo{journal}{Engineering Structures} \bibinfo{volume}{78},
  \bibinfo{pages}{133--144}.
\bibitem[{Spiridonakos and Chatzi(2015)}]{spiridonakos2015metamodeling}
\bibinfo{author}{Spiridonakos, M.D.}, \bibinfo{author}{Chatzi, E.N.},
  \bibinfo{year}{2015}.
\newblock \bibinfo{title}{Metamodeling of nonlinear structural systems with
  parametric uncertainty subject to stochastic dynamic excitation}.
\newblock \bibinfo{journal}{Earthquakes and Structures} \bibinfo{volume}{8},
  \bibinfo{pages}{915--934}.
\bibitem[{Srivastava et~al.(2014)Srivastava, Hinton, Krizhevsky, Sutskever and
  Salakhutdinov}]{srivastava2014dropout}
\bibinfo{author}{Srivastava, N.}, \bibinfo{author}{Hinton, G.},
  \bibinfo{author}{Krizhevsky, A.}, \bibinfo{author}{Sutskever, I.},
  \bibinfo{author}{Salakhutdinov, R.}, \bibinfo{year}{2014}.
\newblock \bibinfo{title}{Dropout: a simple way to prevent neural networks from
  overfitting}.
\newblock \bibinfo{journal}{The journal of machine learning research}
  \bibinfo{volume}{15}, \bibinfo{pages}{1929--1958}.
\bibitem[{Straub and Der~Kiureghian(2008)}]{straub2008improved}
\bibinfo{author}{Straub, D.}, \bibinfo{author}{Der~Kiureghian, A.},
  \bibinfo{year}{2008}.
\newblock \bibinfo{title}{Improved seismic fragility modeling from empirical
  data}.
\newblock \bibinfo{journal}{Structural safety} \bibinfo{volume}{30},
  \bibinfo{pages}{320--336}.
\bibitem[{Tibshirani and Efron(1993)}]{tibshirani1993introduction}
\bibinfo{author}{Tibshirani, R.J.}, \bibinfo{author}{Efron, B.},
  \bibinfo{year}{1993}.
\newblock \bibinfo{title}{An introduction to the bootstrap}.
\newblock \bibinfo{journal}{Monographs on statistics and applied probability}
  \bibinfo{volume}{57}, \bibinfo{pages}{1--436}.
\bibitem[{Towashiraporn(2004)}]{towashiraporn2004building}
\bibinfo{author}{Towashiraporn, P.}, \bibinfo{year}{2004}.
\newblock \bibinfo{title}{Building seismic fragilities using response surface
  metamodels}.
\newblock Ph.D. thesis. Georgia Institute of Technology.
  \bibinfo{address}{Atlanta, GA}.
\bibitem[{{U.S. Geological Survey}(2025)}]{USGS_interactive_hazard}
\bibinfo{author}{{U.S. Geological Survey}}, \bibinfo{year}{2025}.
\newblock \bibinfo{title}{Earthquake hazards program: Interactive map}.
\newblock \URLprefix \url{https://earthquake.usgs.gov/hazards/interactive/}.
  \bibinfo{note}{accessed: 2025-02-19}.
\bibitem[{Vaidyanathan et~al.(2005)Vaidyanathan, Kamatchi and
  Ravichandran}]{vaidyanathan2005artificial}
\bibinfo{author}{Vaidyanathan, C.}, \bibinfo{author}{Kamatchi, P.},
  \bibinfo{author}{Ravichandran, R.}, \bibinfo{year}{2005}.
\newblock \bibinfo{title}{Artificial neural networks for predicting the
  response of structural systems with viscoelastic dampers}.
\newblock \bibinfo{journal}{Computer-Aided Civil and Infrastructure
  Engineering} \bibinfo{volume}{20}, \bibinfo{pages}{294--302}.
\bibitem[{Vamvatsikos and Cornell(2002)}]{vamvatsikos2002incremental}
\bibinfo{author}{Vamvatsikos, D.}, \bibinfo{author}{Cornell, C.A.},
  \bibinfo{year}{2002}.
\newblock \bibinfo{title}{Incremental dynamic analysis}.
\newblock \bibinfo{journal}{Earthquake Engineering \& Structural Dynamics}
  \bibinfo{volume}{31}, \bibinfo{pages}{491--514}.
\bibitem[{Wang and Wu(2020)}]{wang2020knowledge}
\bibinfo{author}{Wang, H.}, \bibinfo{author}{Wu, T.}, \bibinfo{year}{2020}.
\newblock \bibinfo{title}{Knowledge-enhanced deep learning for wind-induced
  nonlinear structural dynamic analysis}.
\newblock \bibinfo{journal}{Journal of Structural Engineering}
  \bibinfo{volume}{146}, \bibinfo{pages}{04020235}.
\bibitem[{Xiong et~al.(2025)Xiong, Jia, Dong and Guo}]{xiong2025uncertainty}
\bibinfo{author}{Xiong, Z.}, \bibinfo{author}{Jia, G.}, \bibinfo{author}{Dong,
  Y.}, \bibinfo{author}{Guo, Y.}, \bibinfo{year}{2025}.
\newblock \bibinfo{title}{Uncertainty-aware fragility modeling of urban
  building exteriors subjected to hurricane-induced windborne debris with
  conditional generative adversarial nets}.
\newblock \bibinfo{journal}{Advances in Wind Engineering} \bibinfo{volume}{2},
  \bibinfo{pages}{100042}.
\bibitem[{Xu and Gardoni(2016)}]{xu2016probabilistic}
\bibinfo{author}{Xu, H.}, \bibinfo{author}{Gardoni, P.}, \bibinfo{year}{2016}.
\newblock \bibinfo{title}{Probabilistic capacity and seismic demand models and
  fragility estimates for reinforced concrete buildings based on
  three-dimensional analyses}.
\newblock \bibinfo{journal}{Engineering Structures} \bibinfo{volume}{112},
  \bibinfo{pages}{200--214}.
\bibitem[{Yang et~al.(2009)Yang, Moehle, Stojadinovic and
  Der~Kiureghian}]{yang2009seismic}
\bibinfo{author}{Yang, T.}, \bibinfo{author}{Moehle, J.},
  \bibinfo{author}{Stojadinovic, B.}, \bibinfo{author}{Der~Kiureghian, A.},
  \bibinfo{year}{2009}.
\newblock \bibinfo{title}{Seismic performance evaluation of facilities:
  Methodology and implementation}.
\newblock \bibinfo{journal}{Journal of Structural Engineering}
  \bibinfo{volume}{135}, \bibinfo{pages}{1146--1154}.
\bibitem[{Zhang et~al.(2020a)Zhang, Liu and Sun}]{zhang2020physicscnn}
\bibinfo{author}{Zhang, R.}, \bibinfo{author}{Liu, Y.}, \bibinfo{author}{Sun,
  H.}, \bibinfo{year}{2020}a.
\newblock \bibinfo{title}{Physics-guided convolutional neural network (phycnn)
  for data-driven seismic response modeling}.
\newblock \bibinfo{journal}{Engineering Structures} \bibinfo{volume}{215},
  \bibinfo{pages}{110704}.
\bibitem[{Zhang et~al.(2020b)Zhang, Liu and Sun}]{zhang2020physics}
\bibinfo{author}{Zhang, R.}, \bibinfo{author}{Liu, Y.}, \bibinfo{author}{Sun,
  H.}, \bibinfo{year}{2020}b.
\newblock \bibinfo{title}{Physics-informed multi-lstm networks for metamodeling
  of nonlinear structures}.
\newblock \bibinfo{journal}{arXiv preprint arXiv:2002.10253} .

\end{thebibliography}



\end{document}